\documentclass[letterpaper]{article} 
\usepackage{aaai25}  
\usepackage{times}  
\usepackage{helvet}  
\usepackage{courier}  
\usepackage[hyphens]{url}  
\usepackage{graphicx} 
\usepackage{natbib}  
\usepackage{caption} 
\usepackage{xcolor}
\usepackage{colortbl}
\usepackage{algorithm}
\usepackage{algorithmic}
\usepackage{amsmath}
\usepackage{amsfonts}
\usepackage{relsize}
\usepackage{booktabs}
\usepackage{multirow}
\usepackage{pifont}
\usepackage{float}
\usepackage{bm}
\newcommand{\cmark}{\ding{51}}%
\newcommand{\xmark}{\ding{55}}%

\usepackage{hyperref}
\hypersetup{
    colorlinks=true,
    linkcolor=red,
    citecolor=black,
    pdfborder={0 0 0}
}

\usepackage{newfloat}
\usepackage{listings}
\DeclareCaptionStyle{ruled}{labelfont=normalfont,labelsep=colon,strut=off} 
\floatstyle{ruled}
\newfloat{listing}{tb}{lst}{}
\floatname{listing}{Listing}

\title{Driving in the Occupancy World: Vision-Centric 4D Occupancy Forecasting \\ and Planning via World Models for Autonomous Driving}
\author{
    Yu Yang \textsuperscript{\rm 1}\equalcontrib,
    Jianbiao Mei \textsuperscript{\rm 1}\equalcontrib,
    Yukai Ma \textsuperscript{\rm 1},
    Siliang Du \textsuperscript{\rm 2 $\dagger$},
    Wenqing Chen \textsuperscript{\rm 2}, \\
    Yijie Qian \textsuperscript{\rm 1}, 
    Yuxiang Feng \textsuperscript{\rm 1},
    Yong Liu \textsuperscript{\rm 1}\thanks{Corresponding authors.}
}
\affiliations{
    \textsuperscript{\rm 1}Zhejiang University \\
    \textsuperscript{\rm 2}Huawei Technologies \\
    \{yu.yang, jianbiaomei, yukaima, yijieqian, yuxiangfeng\}@zju.edu.cn\\
    \{dusiliang, chenwenqing7\}@huawei.com,yongliu@iipc.zju.edu.cn \\
    Project Page: \url{https://drive-occworld.github.io/}
}

\usepackage{bibentry}

\begin{document}


\addtocontents{toc}{\protect\setcounter{tocdepth}{-10}}
\addtocontents{toc}{\protect\contentsline{section}{}{}{}}

\maketitle

\begin{abstract}
World models envision potential future states based on various ego actions. They embed extensive knowledge about the driving environment, facilitating safe and scalable autonomous driving. Most existing methods primarily focus on either data generation or the pretraining paradigms of world models. Unlike the aforementioned prior works, we propose \textbf{Drive-OccWorld}, which adapts a vision-centric 4D forecasting world model to end-to-end planning for autonomous driving. Specifically, we first introduce a semantic and motion-conditional normalization in the memory module, which accumulates semantic and dynamic information from historical BEV embeddings. These BEV features are then conveyed to the world decoder for future occupancy and flow forecasting, considering both geometry and spatiotemporal modeling. Additionally, we propose injecting flexible action conditions, such as velocity, steering angle, trajectory, and commands, into the world model to enable controllable generation and facilitate a broader range of downstream applications. Furthermore, we explore integrating the generative capabilities of the 4D world model with end-to-end planning, enabling continuous forecasting of future states and the selection of optimal trajectories using an occupancy-based cost function. Comprehensive experiments conducted on the nuScenes, nuScenes-Occupancy, and Lyft-Level5 datasets illustrate that our method can generate plausible and controllable 4D occupancy, paving the way for advancements in driving world generation and end-to-end planning.
\end{abstract}

\begin{figure*}[t]
\centering
	\includegraphics[width=0.95\textwidth]{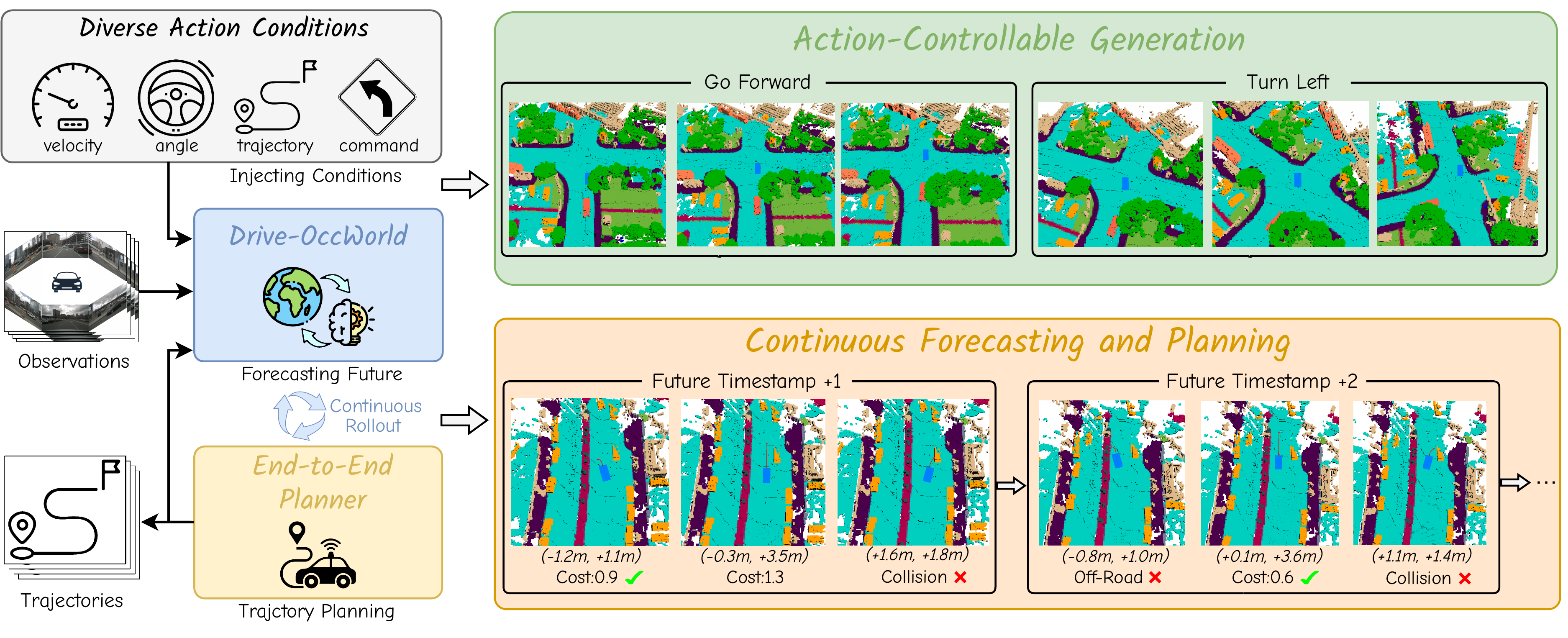}
    \vspace{-1em}
	\caption{\textbf{4D Occupancy Forecasting and Planning via World Model}. Drive-OccWorld takes observations and trajectories as input, incorporating flexible action conditions for \emph{action-controllable generation}. By leveraging world knowledge and the generative capacity of the world model, we further integrate it with a planner for \emph{continuous forecasting and planning}.}
	\label{fig:teaser}
    \vspace{-1em}
\end{figure*}

\section{Introduction}
Autonomous driving (AD) algorithms have advanced significantly in recent decades~\cite{ayoub2019manual, chen2023milestones}. These advancements have transitioned from modular pipelines~\cite{guo2023scenedm, li2023logonet} to end-to-end models~\cite{hu2023planning, jiang2023vad}, which plan trajectories directly from raw sensor data in a unified pipeline. However, due to insufficient world knowledge for forecasting dynamic environments, these methods exhibit deficiencies in generalization ability and safety robustness.

On the other hand, to embed world knowledge and simulate the real-world physics of the driving environment, recent works \cite{zhang2023learning, min2024driveworld, yang2024visual} have introduced the world model \cite{ha2018world} to facilitate scalable autonomous driving. Nevertheless, most of them primarily focus on either data generation or the pretraining paradigms of world models, neglecting the enhancement of safety and robustness for end-to-end planning. For example, many studies \cite{ma2024unleashing, wang2023drivedreamer, hu2023gaia} aimed to generate high-fidelity driving videos through world models to provide additional data for downstream training. The very recent ViDAR \cite{yang2024visual} pre-trained the visual encoder by forecasting point clouds from historical visual input, enhancing performance on downstream tasks such as vision-centric 3D detection and segmentation. Therefore, we believe that integrating the future forecasting capabilities of world models with end-to-end planning remains a worthwhile area for exploration.

In this work, we investigate 4D forecasting and planning using world models to implement future state prediction and end-to-end planning. With the capability to envision various futures based on different ego actions, a world model allows the agent to anticipate potential outcomes in advance. As illustrated in Figure \ref{fig:teaser}, the world model predicts the future state of the environment under different action conditions, using historical observations and various ego actions. Subsequently, the planner employs a cost function that considers both safety and the 3D structure of the environment to select the most suitable trajectory, enabling the agent to navigate effectively in diverse situations. Finally, the predicted future state and selected optimal trajectory can be reintroduced into the world model for the next rollout, facilitating continuous future prediction and trajectory planning. We experimentally demonstrate that leveraging the future forecasting capability of world models enhances the planner's generalization and safety robustness while providing more explainable decision-making, as detailed in Section \ref{sec:exp}.

Specifically, we propose \textbf{Drive-OccWorld}, a vision-centric 4D forecasting and planning world model for autonomous driving. 
Our Drive-OccWorld exhibits three key features: 
\textbf{(1)} \emph{Understanding how the world evolves through 4D occupancy forecasting.} Drive-OccWorld predicts plausible future states based on accumulated historical experiences. It comprises three key components: a history encoder that encodes multi-view geometry BEV embeddings, a memory queue that accumulates historical information, and a future decoder that forecasts occupancy and flows through spatiotemporal modeling. Additionally, we introduce a semantic- and motion-conditional normalization to aggregate significant features.
\textbf{(2)} \emph{Generating various future states based on action conditions.} We incorporate a flexible set of action conditions (\emph{e.g.}, velocity, steering angle, trajectory, and high-level commands), which are encoded and injected into the world decoder through a unified interface, empowering the world model's capability for action-controllable generation.
\textbf{(3)} \emph{Planning trajectories with the world model.} Since the world model can forecast future occupancy and flow, providing perception and prediction results that include the fine-grained states of both agents and background elements, we further design a planner to select the optimal trajectory based on a comprehensive occupancy-based cost function.

We evaluate Drive-OccWorld using the nuScenes, nuScenes-Occupancy, and Lyft-Level5 datasets. 
It outperforms previous methods by 9.5\% in $\text{mIoU}_f$ and 5.1\% in $\text{VPQ}_f$ on nuScenes, by 6.1\% in $\text{mIoU}_f$ and 5.2\% in $\text{VPQ}_f$ on Lyft-Level5 for forecasting the inflated occupancy of movable objects and their 3D backward centripetal flow. 
On the nuScenes-Occupancy benchmark, it achieves a 4.3\% improvement in fine-grained occupancy forecasting. Additionally, experiments on trajectory planning indicate that Drive-OccWorld is effective for safe motion planning.

Our main contributions can be summarized as follows:
\begin{itemize}
    \item We propose Drive-OccWorld, a vision-centric world model designed for forecasting 4D occupancy and dynamic flow, achieving new state-of-the-art performance on both the nuScenes and Lyft benchmarks.
    \item We develop a simple yet efficient semantic- and motion-conditional normalization module for semantic enhancement and motion compensation, which improves forecasting and planning performance.
    \item We incorporate flexible action conditions into Drive-OccWorld to enable action-controllable generation and explore integrating the world model with an occupancy-based planner for continuous forecasting and planning.
\end{itemize}

\begin{figure*}[t]
\centering
	\includegraphics[width=0.98\textwidth]{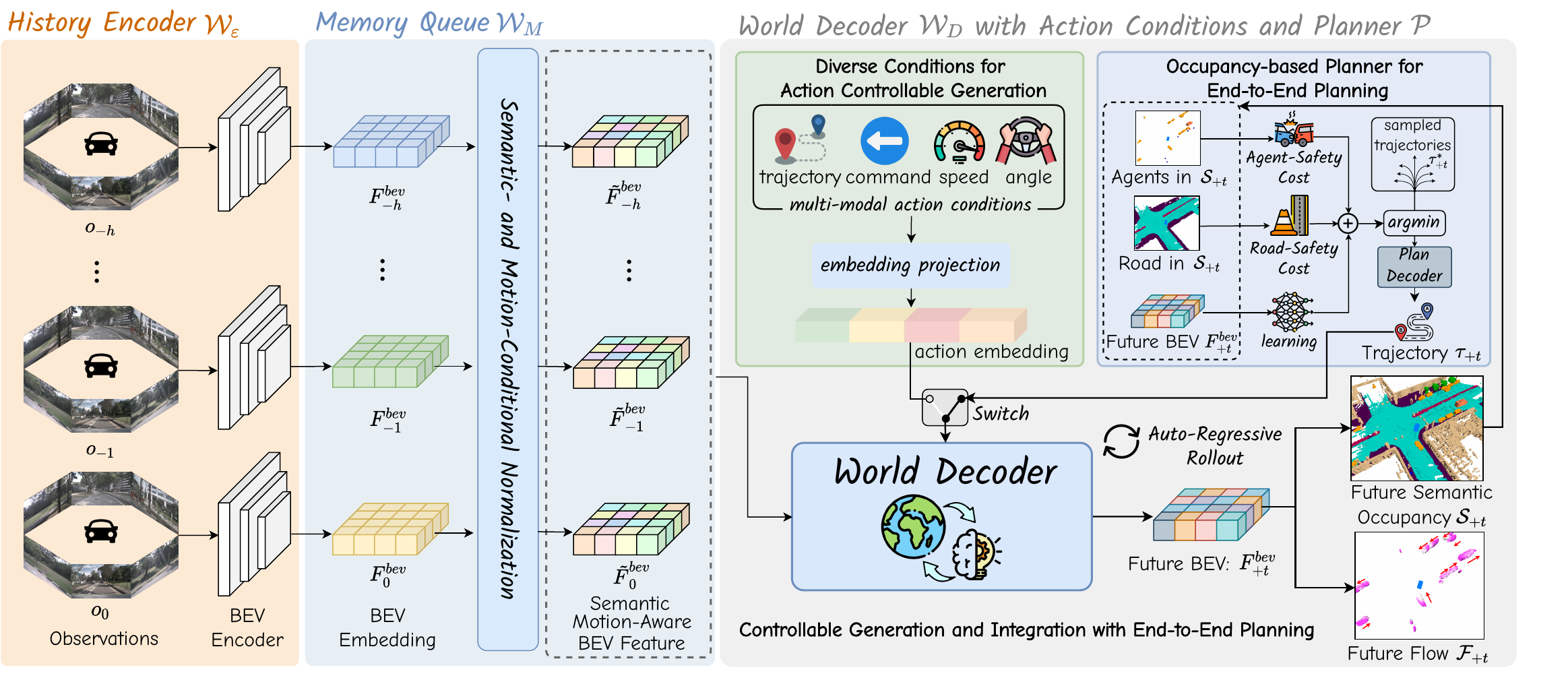}
    \vspace{-.5em}
	\caption{\textbf{Overview of Drive-OccWorld}. (a) The \emph{history encoder} extracts multi-view image features and transforms them into BEV embeddings. (b) The \emph{memory queue} employs semantic- and motion-conditional normalization to aggregate historical information. (c) The \emph{world decoder} incorporates action conditions to generate various future occupancies and flows. Integrating the world decoder with an occupancy-based planner enables continuous forecasting and planning.}
	\label{fig:pipeline}
    \vspace{-1em}
\end{figure*}

\section{Related Works}
\subsection{World Models for Autonomous Driving} 
Existing world models for autonomous driving can be primarily classified into 2D image-based and 3D volume-based models, based on the generation modality of future states.

\paragraph{2D Image-based Models} aim to predict future driving videos using reference images and other conditions (\emph{e.g.}, actions, HDMaps, 3D boxes, and text prompts). GAIA-1 \cite{hu2023gaia} uses an autoregressive transformer as a world model to predict future image tokens based on past image, text, and action tokens. Other methods, such as DriveDreamer \cite{wang2023drivedreamer}, ADriver-I \cite{jia2023adriver}, DrivingDiffusion \cite{li2023drivingdiffusion}, GenAD \cite{yang2024generalized}, Vista \cite{gao2024vista}, Delphi \cite{ma2024unleashing}, and Drive-WM \cite{wang2024driving}, use \emph{latent diffusion models} (LDMs) \cite{rombach2022high, blattmann2023stable} for image-to-driving video generation. These methods focus on designing modules to incorporate actions, BEV layouts, and other priors into the denoising process, resulting in more coherent and plausible future video generations.

\paragraph{3D Volume-based Models} forecast future states in the form of point clouds or occupancy. Copilot4D \cite{zhang2023learning} tokenizes LiDAR observations with VQVAE \cite{van2017neural} and predicts future point clouds via discrete diffusion. ViDAR \cite{yang2024visual} implements a visual point cloud forecasting task to pre-train visual encoders. UnO \cite{agro2024uno} forecasts a continuous occupancy field with self-supervision from LiDAR data. OccWorld \cite{zheng2023occworld} and OccSora \cite{wang2024occsora} compact the occupancy input with a scene tokenizer and use a generative transformer to predict future occupancy. UniWorld \cite{min2023uniworld} and DriveWorld \cite{min2024driveworld} propose 4D pre-training via 4D occupancy reconstruction.

In this work, we investigate potential applications of the world model by injecting action conditions to enable action-controllable generation and integrating this generative capability with end-to-end planners for safe driving.

\section{Method}
\subsection{Preliminary}
An end-to-end autonomous driving model aims to control a vehicle (\emph{i.e.}, plan trajectories) directly based on sensor inputs and ego actions \cite{hu2023planning}. Formally, given historical sensor observations $\{o_{-h}, \ldots, o_{-1}, o_{0}\}$ and ego trajectories $\{\tau_{-h}, \ldots, \tau_{-1}, \tau_{0}\}$ over $h$ timestamps, an end-to-end model $\mathcal{A}$ predicts desirable ego trajectories $\{\tau_{1}, \ldots, \tau_{f}\}$ for the future $f$ timestamps:
\begin{equation}
    \mathcal{A}( \{o_{-h}, \ldots, o_{-1}, o_{0}\}, \{\tau_{-h}, \ldots, \tau_{-1}, \tau_{0}\} ) = \{\tau_{1}, \ldots, \tau_{f}\} 
\end{equation}
    
A driving world model $\mathcal{W}$ can be viewed as a generative model that takes prior observations and ego actions $\{a_{-h}, \ldots, a_{-1}, a_{0}\}$ as input, generating plausible future states $\{s_{1}, \ldots, s_{f}\}$ of the environment:
\begin{equation}
    \mathcal{W}( \{o_{-h}, \ldots, o_{-1}, o_{0}\}, \{a_{-h}, \ldots, a_{-1}, a_{0}\} ) = \{s_{1}, \ldots, s_{f}\} 
\end{equation}
where ego actions $a$ can be injected into the controllable generation process in various forms, \emph{i.e.}, velocity, steering angle, ego trajectory, and high-level commands.

Given the world model's ability to foresee future states, we propose integrating it with a planner to fully exploit the capabilities of the world model in end-to-end planning. Specifically, we introduce an auto-regressive framework termed Drive-OccWorld, which consists of a generative world model $\mathcal{W}$ to forecast future occupancy and flow states, and an occupancy-based planner $\mathcal{P}$ that employs a cost function to select the optimal trajectory based on evaluating future predictions. Formally, we formulate Drive-OccWorld as follows, which auto-regressively predicts the future state and trajectory at the next timestamp:
\begin{equation}
\begin{split}
    \mathcal{W}( & \{o_{-h}, \ldots, o_{-1}, o_{0}\}, \{s_{1}, \ldots, s_{t-1}, s_{t}\}, \\
    & \{a_{-h}, \ldots, a_{-1}, a_{0}, \ldots, a_{t-1}, a_{t}\} ) = s_{t+1}
\end{split}
\label{eq2}
\end{equation}
\begin{equation}
     \mathcal{P} ( f_{o}(s_{t+1}, \tau_{t+1}^*)) = \tau_{t+1}
     \label{eq3}
\end{equation}
where $f_o$ is the occupancy-based cost function, and $\tau_{t+1}^*$ denotes sampled trajectory proposals at the $t+1$ timestamp. 

Notably, for action-controllable generation, $a$ can be injected into $\mathcal{W}$ as conditions in the form of velocity, \emph{etc.}, and $\mathcal{P}$ is discarded to prevent potential ego-status leakage. In end-to-end planning, the predicted trajectory $\tau_{t+1}$ serves as the action condition $a_{t+1}$ for forecasting the next state $s_{t+2}$, leading to a continuous rollout of forecasting and planning.

In the following sections, we will detail the world model's structure, equipping $\mathcal{W}$ with action-controllable generation and integrating it with $\mathcal{P}$ for end-to-end planning.

\subsection{4D Forecasting with World Model}
As depicted in Figure \ref{fig:pipeline}, Drive-OccWorld comprises three components: \textbf{(1)} a \emph{History Encoder} $\mathcal{W}_\mathcal{E}$, which takes historical camera images as input, extracts multi-view geometry features, and transforms them into BEV embeddings. Following previous works \cite{yang2024visual, min2024driveworld}, we utilize the visual BEV encoder \cite{li2022bevformer} as our history encoder. \textbf{(2)} a \emph{Memory Queue} $\mathcal{W}_\mathcal{M}$ \emph{with Semantic- and Motion-Conditional Normalization}, which employs a simple yet efficient normalization operation in latent space to aggregate semantic information and compensate for dynamic motions, thereby accumulating more representative BEV features. \textbf{(3)} a \emph{World Decoder} $\mathcal{W}_\mathcal{D}$, which extracts world knowledge through temporal modeling with historical features to forecast future semantic occupancies and flows. Flexible action conditions can be injected into $\mathcal{W}_\mathcal{D}$ for controllable generation. An occupancy-based planner $\mathcal{P}$ is integrated for continuous forecasting and planning.

\begin{figure}[t]
\centering
	\includegraphics[width=0.98\linewidth]{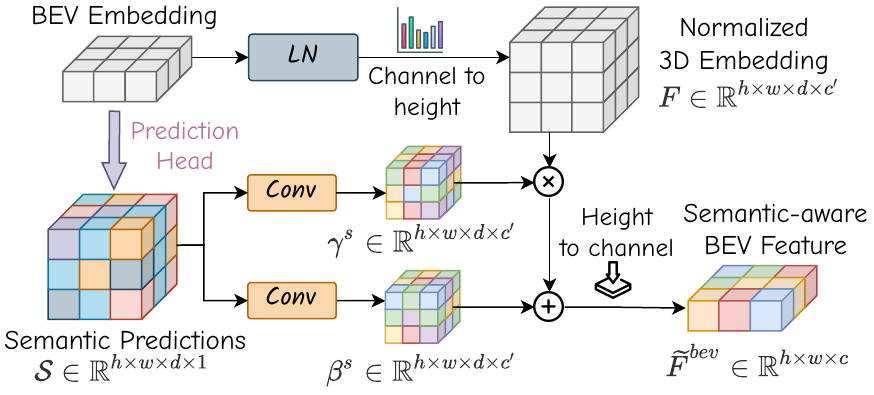}
	\vspace{-1em}
    \caption{Overview of semantic-conditional normalization.}
	\label{fig:norm}
    \vspace{-1em}
\end{figure}

\paragraph{Semantic- and Motion-Conditional Normalization} \label{norm}
is designed to enhance historical BEV embeddings by incorporating semantic and dynamic information. For example, consider the BEV embedding $\bm{F}^{bev} \in \mathbb{R}^{h \times w \times c}$, where $h$ and $w$ are the spatial resolutions of the BEV, and $c$ denotes the channel dimension. We first apply layer normalization without affine mapping, then modulate it into $\tilde{\bm{F}}^{bev}$ using an adaptive affine transformation, with the scale and shift parameters $(\gamma^*, \beta^*)$ derived from semantic or motion labels:
\begin{equation}
    \tilde{\bm{F}}^{bev} = \gamma^* \cdot \textit{LayerNorm}(\bm{F}^{bev}) + \beta^*
    \label{eq_norm}
\end{equation}

Specifically, for semantic-conditional normalization, $(\gamma^s, \beta^s)$ are inferred from voxel-wise semantic predictions. As illustrated in Figure~\ref{fig:norm}, we implement a lightweight head along with the argmax function to predict voxel-wise semantic labels $\mathcal{S} \in \mathbb{R}^{h \times w \times d \times 1}$, where $d$ denotes the height of the voxelized 3D space. The semantic labels are encoded as one-hot embeddings and convolved to produce modulation parameters for the affine transformation as Eq.~\ref{eq_norm}. This method efficiently enhances the semantic discrimination of BEV embeddings, as demonstrated in the experiments.

In motion-conditional normalization, we account for the movements of both the ego vehicle and other agents across various timestamps. Specifically, the ego-pose transformation matrix $E_{-t}^{+t} = [R_{-t}^{+t}, T_{-t}^{+t}]$, which represents the rotation and translation of the ego vehicle from timestamp $-t$ to $+t$, is flattened and encoded into an embedding processed by MLPs to generate affine transformation parameters $(\gamma^e, \beta^e)$. To address the movements of other agents, we predict voxel-wise 3D backward centripetal flow $\mathcal{F} \in \mathbb{R}^{h \times w \times d \times 3}$ that points from the voxel at time $t$ to its corresponding 3D instance center at $t-1$, and encode it into $(\gamma^f, \beta^f)$ for fine-grained motion-aware normalization using Eq.~\ref{eq_norm}.

\paragraph{Future Forecasting with World Decoder.} $\mathcal{W}_\mathcal{D}$ is an auto-regressive transformer that predicts the BEV embeddings $\bm{F}_{+t}^{bev}$ for the future frame $+t$ based on historical BEV features stored in $\mathcal{W}_\mathcal{M}$ and the expected action condition $a_{+t}$.

Specifically, $\mathcal{W}_\mathcal{D}$ takes learnable BEV queries as input and performs deformable self-attention, temporal cross-attention with historical embeddings, conditional cross-attention with action conditions, and a feedforward network to generate future BEV embeddings. 
The conditional layer performs cross-attention between BEV queries and action embeddings, which will be illustrated in the following section, injecting action-controllable information into the forecasting process. After obtaining the next BEV embeddings $\bm{F}_{+t}^{bev}$, prediction heads utilizing the channel-to-height operation \cite{yu2023flashocc} to predict semantic occupancy and 3D backward centripetal flow $(\mathcal{S}_{+t}, \mathcal{F}_{+t}) \in \mathbb{R}^{h \times w \times d}$.

In the training process, we employ multiple losses, including cross-entropy loss, Lovász loss \cite{berman2018lovasz}, and binary occupancy loss, to constrain the semantics and geometries of occupancy predictions $\mathcal{S}_{1:f}$. The $l_1$ loss is used to supervise flow predictions $\mathcal{F}_{1:f}$.

\subsection{Action-Controllable Generation}
\label{action}
Due to the inherent complexity of the real world, the motion states of the ego vehicle are crucial for the world model to understand how the agent interacts with its environment. Therefore, to fully comprehend the environment, we propose leveraging diverse action conditions to empower Drive-OccWorld with the capability for controllable generation.

\paragraph{Diverse Action Conditions} include multiple formats: \textbf{(1) \emph{Velocity}} is defined at a given time step as $(v_x, v_y)$, representing the speeds of the ego vehicle decomposed along the $x$ and $y$ axes in $m/s$. \textbf{(2) \emph{Steering Angle}} is collected from the steering feedback sensor. Following VAD, we convert it into curvature in $m^{-1}$, indicating the reciprocal of the turning radius while considering the geometric structure of the ego car. \textbf{(3) \emph{Trajectory}} represents the movement of the ego vehicle's location to the next timestamp, formulated as $(\triangle x, \triangle y)$ in meters. It is widely used as the output of end-to-end methods, including our planner $\mathcal{P}$. \textbf{(4) \emph{Commands}} consist of go forward, turn left, and turn right, which represent the highest-level intentions for controlling the vehicle.

\paragraph{Unified Conditioning Interface} is designed to incorporate heterogeneous action conditions into a coherent embedding, inspired by \cite{gao2024vista, wang2024driving}. We first encode the required actions via Fourier embeddings \cite{tancik2020fourier}, which are then concatenated and fused via learned projections to align with the dimensions of the conditional cross-attention layers in $\mathcal{W}_\mathcal{D}$. This method enables efficient integration of flexible action conditions into controllable generation, with experiments demonstrating that the unified interface with conditional cross-attention provides superior controllability compared to other approaches such as additive embeddings.

\subsection{End-to-End Planning with World Model}
Existing world models primarily focus on either data generation or the pertaining paradigms for autonomous driving. Although a recent pioneering work, Drive-WM~\cite{wang2024driving}, proposed integrating generated driving videos with an image-based reward function to plan trajectories, the geometric 3D features of the environment are not fully exploited for motion planning.
Leveraging the future occupancy forecasting capabilities of our world model, as illustrated in Figure~\ref{fig:pipeline}, we introduce an occupancy-based planner that samples occupied grids of agents and drivable areas to enforce safety constraints. Additionally, we employ a learned-volume cost to provide a more comprehensive evaluation of the environment for safe planning.

\paragraph{Occupancy-based Cost Function} is designed to ensure the safe driving of the ego vehicle. It consists of multiple cost factors: \textbf{(1) \emph{Agent-Safety Cost}} constrains the ego vehicle from colliding with other agents, such as pedestrians and vehicles. It penalizes trajectory candidates that overlap with grids occupied by other road users. Additionally, trajectories that are too close to other agents, in terms of lateral or longitudinal distance, are also restricted to avoid potential collisions. \textbf{(2) \emph{Road-Safety Cost}} ensures the vehicle remains on the road. It extracts road layouts from occupancy predictions, penalizing trajectories that fall outside the drivable area. \textbf{(3) \emph{Learned-Volume Cost}} is inspired by ST-P3 \cite{hu2022st}. It employs a learnable head based on $\bm{F}_{+t}^{bev}$ to generate a 2D cost map, enabling a more comprehensive evaluation of occupancy grids in complex environments.

The total cost function is the summation of the above cost factors. Following the approach of ST-P3, a trajectory sampler generates a set of candidate trajectories $\tau_{+t}^* \in \mathbb{R}^{N_{\tau} \times 2}$ distributed across the 2D grid map surrounding the ego vehicle, guided by high-level commands. Subsequently, the trajectory planner $\mathcal{P}$ selects the optimal trajectory $\tau_{+t}$ by minimizing the total cost function, while simultaneously ensuring agent and road safety.

\paragraph{BEV Refinement} is introduced to further refine the trajectory using the latent features of BEV embeddings $\bm{F}_{+t}^{bev}$. We encode $\tau_{+t}$ into an embedding and concatenate it with a command embedding to form an ego query, which performs cross-attention with $\bm{F}_{+t}^{bev}$ to extract fine-grained representations of the environment. The final trajectory is predicted based on the refined ego query through MLPs. 

The planning loss $\mathcal{L}_{plan}$ consists of three components: a max-margin loss introduced by \cite{sadat2020perceive} to constrain the safety of trajectory candidates $\tau_{+t}^*$, a naive $l_2$ loss for imitation learning, and a collision loss that ensures the planned trajectory avoids grids occupied by obstacles.
Notably, when performing end-to-end planning, we utilize the predicted trajectories as action conditions for both training and testing. This approach not only prevents GT ego actions from leaking into the planner but also facilitates model learning using predicted trajectories, enabling improved performance during testing.

\begin{table*}[t]
\fontsize{9}{10}\selectfont
\setlength{\tabcolsep}{1.1pt}
\center
\renewcommand\arraystretch{1.0}
\definecolor{mycolor}{rgb}{0.64, 0.87, 0.93}
\begin{tabular}{l|cccc|cccc|ccc}
\toprule
\multicolumn{1}{l|}{\multirow{3}{*}{Method}} & \multicolumn{8}{c|}{Inflated GMO} & \multicolumn{3}{c}{Fine-Grained GMO} \\
\cmidrule{2-12}
\multicolumn{1}{c|}{} & \multicolumn{4}{c|}{\cellcolor{gray!10}nuScenes} & \multicolumn{4}{c|}{\cellcolor{gray!10}Lyft-Level5} & \multicolumn{3}{c}{\cellcolor{gray!10}nuScenes-Occupancy} \\ 
\multicolumn{1}{c|}{} & $\text{mIoU}_c$ & $\text{mIoU}_f$ (2\,s) & $\tilde{\text{mIoU}}_f$ & $\text{VPQ}_f$ & $\text{IoU}_c$ & $\text{mIoU}_f$ (0.8\,s)  & $\tilde{\text{mIoU}}_f$ & $\text{VPQ}_f$ & $\text{mIoU}_c$ & $\text{mIoU}_f$ (2\,s) & $\tilde{\text{mIoU}}_f$   \\ 
\cmidrule{1-12}
SPC & 1.3 & failed & failed & -- & 1.4 & failed & failed & -- & 5.9 & 1.1 & 1.1  \\ 
OpenOccupancy-C \cite{wang2023openoccupancy} & 12.2 & 11.5 & 11.7 & -- & 14.0 & 13.5 & 13.7 & -- & 10.8 & 8.0 & 8.5 \\
PowerBEV-3D \cite{li2023powerbev} & 23.1 & 21.3 & 21.9 & 20.0 & 26.2 & 24.5 & 25.1 & 27.4 & 5.9 & 5.3 & 5.5 \\ 
Cam4DOcc \cite{ma2024cam4docc} & 31.3 & 26.8 & 28.0 & 18.6 & 36.4 & 33.6 & 34.6 & 28.2 & 11.5 & 9.7 & 10.1 \\ 
\cmidrule{1-12}
\rowcolor{mycolor!25} Drive-OccWorld$^\mathcal{A}$ (Ours) & \underline{39.7} & \textbf{36.3} & \underline{37.3} & \underline{23.7} & \underline{40.6} & \underline{39.3} & \underline{40.0} & \underline{32.2}  & \textbf{13.6} & \underline{11.9} & \underline{12.3} \\ 
\rowcolor{mycolor!25} Drive-OccWorld$^\mathcal{P}$ (Ours) & \textbf{39.8} & \textbf{36.3} & \textbf{37.4} & \textbf{25.1} & \textbf{40.9} & \textbf{39.7} & \textbf{40.6} & \textbf{33.4} & \textbf{13.6} & \textbf{12.0} & \textbf{12.4} \\ 
\bottomrule
\multicolumn{12}{l}{SPC: SurroundDepth \cite{wei2023surrounddepth} + PCPNet \cite{luo2023pcpnet} + Cylinder3D \cite{zhu2021cylindrical}} \\
\end{tabular}
\vspace{-.6em}
\caption{Comparisons of \textbf{Inflated GMO and Flow Forecasting} on the nuScenes and Lyft-Level5 datasets, and \textbf{Fine-Grained GMO Forecasting} on the nuScenes-Occupancy dataset, with the top two results highlighted in bold and underlined text.}
\label{tab1}
\vspace{-.6em}
\end{table*}

\section{Experiments} \label{sec:exp}
\subsection{Setup}
We conduct experiments on the nuScenes \cite{caesar2020nuscenes}, nuScenes-Occupancy \cite{wang2023openoccupancy}, and Lyft-Level5 datasets. More details regarding the experimental implementation are provided in the appendix.

\paragraph{Tasks Definition.}
We validate Drive-OccWorld's effectiveness in 4D occupancy and flow forecasting, which includes inflated and fine-grained occupancy formats, as well as trajectory planning.
\textbf{(1) \emph{Inflated GMO and Flow Forecasting}} is presented in Cam4DOcc \cite{ma2024cam4docc}, predicting the future states of \emph{general movable objects (GMO)} with dilated occupancy patterns. The dataset is constructed by transforming movable objects at different timestamps into the present coordinate system, voxelizing the 3D space, and labeling grids with semantic and instance annotations. Voxel grids within bounding boxes are marked as GMO, and the 3D backward centripetal flow indicates the direction of voxels in the current frame toward their corresponding 3D instance centers from the previous timestamp.
\textbf{(2) \emph{Fine-Grained GMO Forecasting}} predicts occupied grids of GMO using voxel-wise labels from the nuScenes-Occupancy dataset.
\textbf{(3) \emph{Fine-grained GMO and GSO Forecasting}} predicts the voxel-wise labels for both \emph{general movable objects (GMO)} and \emph{general static objects (GSO)} based on fine-grained annotations from nuScenes-Occupancy.
\textbf{(4) \emph{End-to-end Planning}} follows the open-loop evaluation on nuScenes. 

\paragraph{Metrics.}
\textbf{(1) \emph{Occupancy forecasting}} is evaluated using the mIoU metric. Following Cam4DOcc, we assess the current moment $(t=0)$ with $\text{mIoU}_c$ and the future timestamps $(t \in [1,f])$ with $\text{mIoU}_f$. Additionally, we provide a quantitative indicator $\tilde{\text{mIoU}}_f$ weighted by timestamp, in line with the principle that occupancy predictions at nearby timestamps are more critical for planning. 
\textbf{(2) \emph{Flow predictions}} are evaluated through instance association using the video panoptic quality $\text{VPQ}_f$ metric. 
We further report the flow forecasting results denoted as $\text{VPQ}_f^*$, utilizing a simple yet efficient center clustering technique, where the predicted object centers are clustered based on their relative distances.
\textbf{(3) \emph{End-to-end planning}} is evaluated using the L2 distance from ground truth trajectories and the object collision rate.
More details of the metrics are provided in the appendix.

\begin{figure*}[t]
\centering
	\includegraphics[width=0.98\textwidth]{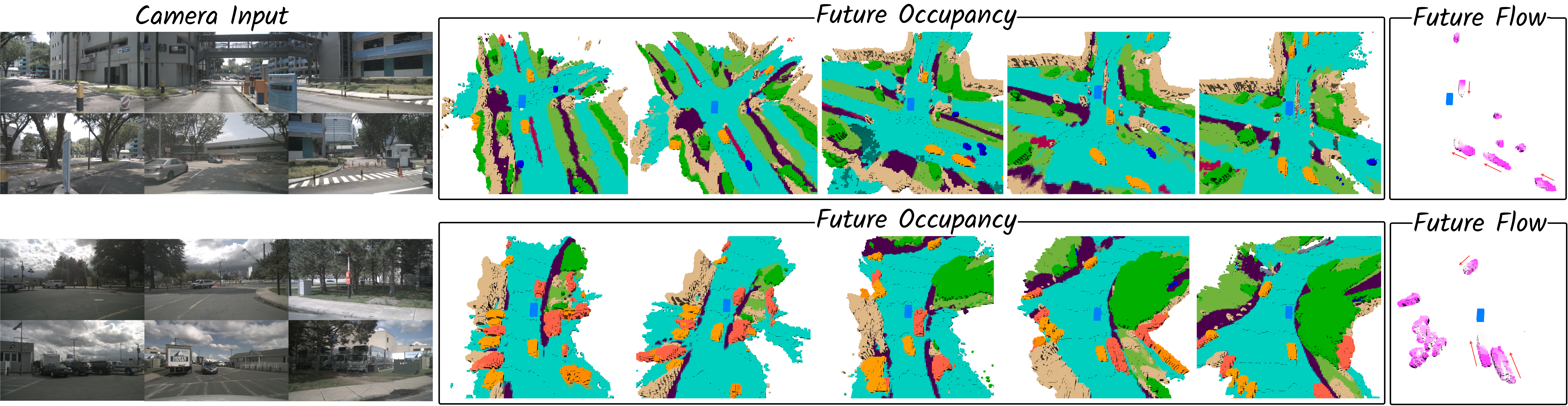}
    \vspace{-.5em}
	\caption{\textbf{Qualitative results of 4D occupancy and flow forecasting}. The results are presented at various future timestamps.}
	\label{fig:quantative}
    \vspace{-1em}
\end{figure*}

\begin{table}[t] 
\fontsize{8}{10}\selectfont
\setlength{\tabcolsep}{0.5mm}
\center
\definecolor{mycolor}{rgb}{0.64, 0.87, 0.93}
\begin{tabular}{l|ccccccc}
\toprule
\multirow{2}{*}{Method} & \multicolumn{3}{c}{$\text{mIoU}_c$} & \multicolumn{3}{c}{$\text{mIoU}_f$ (2\,s)} & \multicolumn{1}{c}{$\tilde{\text{mIoU}}_f$}                                         \\ \cmidrule{2-8} & \rotatebox{60}{GMO} & \rotatebox{60}{GSO} & \rotatebox{60}{mean} & \rotatebox{60}{GMO} & \rotatebox{60}{GSO} & \rotatebox{60}{mean} & \rotatebox{60}{GMO}   \\ \cmidrule{1-8}
SPC & 5.9 & 3.3 & 4.6 & 1.1 & 1.4 & 1.2 & 1.1 \\ 
PowerBEV-3D \cite{li2023powerbev} & 5.9 & -- & -- & 5.3 & -- & -- & 5.5 \\
CONet-C \cite{wang2023openoccupancy} & 9.6 &  17.2 &  13.4 & 7.4 & 17.3 & 12.4 & 7.9 \\ 
Cam4DOcc \cite{ma2024cam4docc} & 11.0  & 17.8 & 14.4 & 9.2  & 17.8  & 13.5 & 9.7  \\  \cmidrule{1-8}
\rowcolor{mycolor!25} DriveOccWorld$^{\mathcal{A}}$ (Ours)  & \underline{16.6} & \textbf{20.2} & \underline{18.4} & \textbf{14.3} & \textbf{21.4} & \textbf{17.8} & \textbf{14.9} \\
\rowcolor{mycolor!25} DriveOccWorld$^{\mathcal{P}}$ (Ours) & \textbf{16.9} & \textbf{20.2} & \textbf{18.5} & \textbf{14.3} & \underline{21.2} & \textbf{17.8} & \textbf{14.9}  \\
\bottomrule
\multicolumn{8}{l}{SPC: SurroundDepth \cite{wei2023surrounddepth} + PCPNet \cite{luo2023pcpnet}} \\
\multicolumn{8}{l}{+ Cylinder3D \cite{zhu2021cylindrical}}
\end{tabular}
\vspace{-.5em}
\caption{Comparisons of \textbf{Fine-Grained GMO and GSO Forecasting} on nuScenes-Occupancy dataset.}
\vspace{-1.5em}
\label{tab2}
\end{table}

\subsection{Main Results of 4D Occupancy Forecasting}
First, we verify the quality of 4D occupancy forecasting and the controllable generation capabilities of Drive-OccWorld. We report performance conditioned on ground-truth actions as Drive-OccWorld$^\mathcal{A}$, and results conditioned on predicted trajectories from the planner $\mathcal{P}$ as Drive-OccWorld$^\mathcal{P}$.

\paragraph{Inflated GMO and Flow Forecasting.}
Table~\ref{tab1} presents comparisons of inflated GMO and flow forecasting on the nuScenes and Lyft-Level5 datasets. Drive-OccWorld outperforms previous methods on various time intervals, including the performance on the current moment and future timestamps. For instance, it surpasses Cam4DOcc on the $\tilde{\text{mIoU}}_f$ metric by 9.4\% and 6\% on nuScenes and Lyft-Level5, respectively. The results demonstrate Drive-OccWorld's superior ability to forecast future world states.

For future flow predictions, Drive-OccWorld$^\mathcal{P}$ outperforms previous SoTA methods by 5.1\% and 5.2\% on $\text{VPQ}_f$ for nuScenes and Lyft-Level5, respectively, indicating superior capability in modeling dynamic object motions.

\paragraph{Fine-Grained GMO Forecasting.}
Fine-grained GMO forecasting poses more challenges than inflated GMO forecasting, since it requires predicting more fine-grained voxel labels than the bounding-box level labels. Table~\ref{tab1} presents comparisons of fine-grained GMO forecasting on the nuScenes-Occupancy dataset. The results show that Drive-OccWorld still outperforms previous methods across all time intervals, showcasing Drive-OccWorld's ability to forecast more fine-grained world states.

\begin{table}[t]
\fontsize{8}{10}\selectfont
\setlength{\tabcolsep}{0.8mm}
\center
\definecolor{mycolor}{rgb}{0.64, 0.87, 0.93}
\begin{tabular}{c|cccc|ccc|c}
\toprule
\multirow{2}{*}{No.} & \multicolumn{4}{c|}{Action Condition} & \multirow{2}{*}{$\text{mIoU}_c$} & \multirow{2}{*}{$\text{mIoU}_f$ (1\,s)} & \multirow{2}{*}{$\tilde{\text{mIoU}}_f$} & \multirow{2}{*}{$\text{VPQ}_f^*$} \\
& traj   & vel   & angle   & cmd   & & & &    \\ \cmidrule{1-9}
1 & & & & & 28.7 & 26.4 & 26.8 & 33.5 \\
2 & \cmark & & & & 28.5 & \underline{27.6}\textcolor{red}{$_{\mathbf{\uparrow}\textbf{1.2}}$} & \underline{27.8}\textcolor{red}{$_{\uparrow\textbf{1.0}}$} & 33.7\textcolor{red}{$_{\uparrow\text{0.2}}$} \\
3 &  & \cmark & & & 28.9\textcolor{red}{$_{\uparrow\text{0.2}}$} & 27.5\textcolor{red}{$_{\uparrow\textbf{1.1}}$} & \underline{27.8}\textcolor{red}{$_{\uparrow\textbf{1.0}}$} & 33.9\textcolor{red}{$_{\uparrow\text{0.4}}$} \\
4 & & & \cmark & & 28.9\textcolor{red}{$_{\uparrow\text{0.2}}$} & 26.8\textcolor{red}{$_{\uparrow\text{0.4}}$} & 27.2\textcolor{red}{$_{\uparrow\text{0.4}}$} & 34.2\textcolor{red}{$_{\uparrow\text{0.7}}$} \\
5 &  & & & \cmark & \textbf{29.2}\textcolor{red}{$_{\uparrow\textbf{0.5}}$} & 26.8\textcolor{red}{$_{\uparrow\text{0.4}}$} & 27.3\textcolor{red}{$_{\uparrow\text{0.5}}$} & 34.7\textcolor{red}{$_{\uparrow\text{1.2}}$} \\
6 & \cmark & \cmark & \cmark & \cmark & \underline{29.0}\textcolor{red}{$_{\uparrow\text{0.3}}$} & \underline{27.6}\textcolor{red}{$_{\uparrow\textbf{1.2}}$} & \underline{27.8}\textcolor{red}{$_{\uparrow\textbf{1.0}}$} & \underline{35.0}\textcolor{red}{$_{\uparrow\text{1.5}}$} \\ \cmidrule{1-9}
\rowcolor{mycolor!25} 7 & \cmark$^{\mathcal{P}}$ & & & & \textbf{29.2}\textcolor{red}{$_{\uparrow\textbf{0.5}}$} & \textbf{27.9}\textcolor{red}{$_{\uparrow\textbf{1.5}}$} & \textbf{28.1}\textcolor{red}{$_{\uparrow\textbf{1.3}}$} & \textbf{35.1}\textcolor{red}{$_{\uparrow\textbf{1.6}}$} \\ 
\bottomrule
\end{tabular}
\vspace{-.5em}
\caption{\textbf{Comparisons of controllability under diverse action conditions}, with the top two results highlighted in bold and underlined. \cmark$^{\mathcal{P}}$ denotes the predicted trajectory.}
\label{tab3}
\vspace{-1.5em}
\end{table}

\paragraph{Fine-grained GMO and GSO Forecasting.}
Table~\ref{tab2} presents comparisons of fine-grained GMO and GSO forecasting on the nuScenes-Occupancy dataset. The results demonstrate that Drive-OccWorld achieves the best performance among all approaches. Notably, Drive-OccWorld$^\mathcal{P}$ outperforms Cam4DOcc by 5.9\% and 5.1\% on $\text{mIoU}$ for general movable objects (GMO) at current and future timestamps, respectively, illustrating its ability to accurately locate movable objects for safe navigation. Figure~\ref{fig:quantative} provides qualitative results of fine-grained occupancy forecasting and flow predictions across frames.

\begin{figure*}[t]
\centering
	\includegraphics[width=0.98\textwidth]{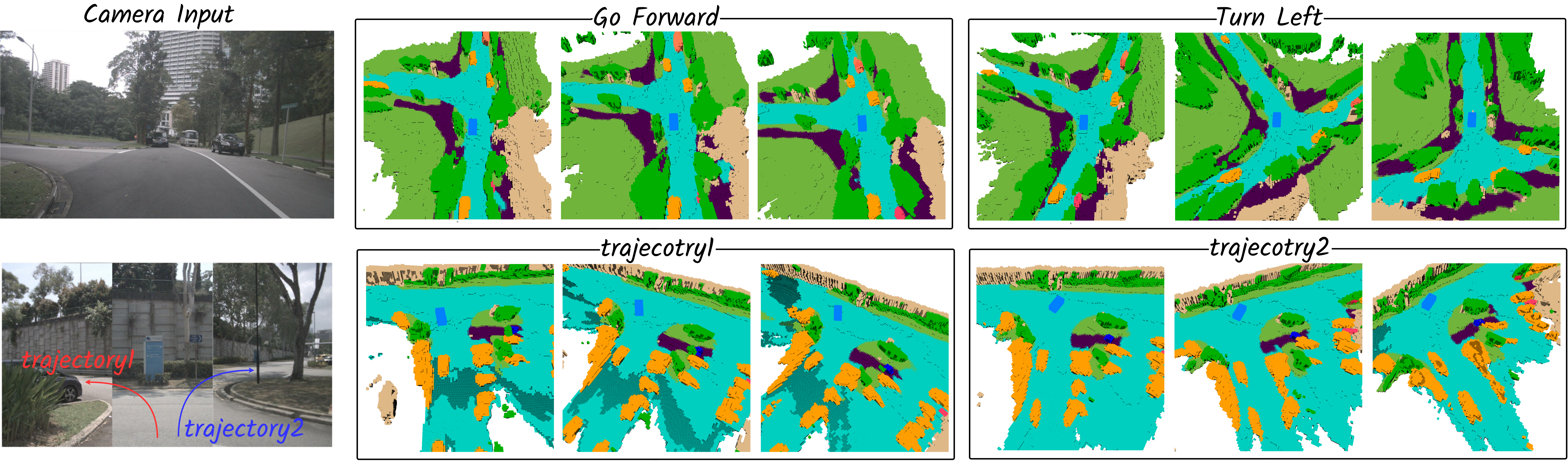}
    \vspace{-.5em}
	\caption{\textbf{Qualitative results of controllable generation}, using the high-level \emph{command} or low-level \emph{trajectory} conditions.}
	\label{fig:contorl}
    \vspace{-1em}
\end{figure*}

\paragraph{Controllability.}
Table~\ref{tab3} examines controllability under various action conditions. Injecting any action condition improves results compared to the baseline. Low-level conditions, such as trajectory and velocity, significantly enhance future forecasting ($\tilde{\text{mIoU}}_f$), while high-level conditions, such as commands, boost results at the current moment ($\text{mIoU}_c$) but have limited impact on future predictions. We explain that incorporating more low-level conditions helps the world model better understand how the ego vehicle interacts with the environment, thereby improving forecasting performance.

\begin{table}[t]
\scriptsize
\setlength{\tabcolsep}{3.0pt}
\center
\renewcommand\arraystretch{0.9}
\definecolor{mycolor}{rgb}{0.64, 0.87, 0.93}
\scriptsize
\begin{tabular}{c|cccc|cccc}
\toprule
\multirow{2}{*}{Action Condition}  & \multicolumn{4}{c|}{L2 (m) $\downarrow$} & \multicolumn{4}{c}{Collision (\%) $\downarrow$} \\
 & {1s} & {2s} & {3s} & \cellcolor{gray!10}{Avg.} & {1s} & {2s} & {3s} & \cellcolor{gray!10}{Avg.} \\
\midrule
\textcolor{gray}{\textbf{GT trajectory}} & \textcolor{gray}{\textbf{0.26}} & \textcolor{gray}{\textbf{0.52}} & \textcolor{gray}{\textbf{0.89}} & \cellcolor{gray!10}\textcolor{gray}{\textbf{0.56}} & \textcolor{gray}{\textbf{0.02}} & \textcolor{gray}{\textbf{0.11}} & \textcolor{gray}{\textbf{0.36}} & \cellcolor{gray!10}\textcolor{gray}{\textbf{0.16}}\\
\rowcolor{mycolor!25} Pred trajectory & 0.32 & 0.75 & 1.49 & \cellcolor{gray!10}0.85 & 0.05 & 0.17 & 0.64 & \cellcolor{gray!10}0.29\\
\bottomrule
\end{tabular}
\captionsetup{aboveskip=1.0em}
\setlength{\belowcaptionskip}{-2.3em}
\caption{\textbf{Planning upper bound} when using GT trajectory.}
\label{tab8}
\end{table}

Table~\ref{tab8} shows that using ground-truth trajectories as action conditions yields better planning results than predicted trajectories. However, using predicted trajectories slightly improves occupancy and flow forecasting quality, as indicated by comparisons in Table~\ref{tab3} (line 2 vs. line 7) and supported by Tables~\ref{tab1} and \ref{tab2}, where Drive-OccWorld$^\mathcal{P}$ outperforms Drive-OccWorld$^\mathcal{A}$. This performance gain may stem from planning constraints associated with predicted trajectories, allowing the planner to perform cross-attention between trajectories and BEV features. This process constrains the BEV features to account for ego motion, thereby enhancing perception performance. Additionally, using predicted trajectories during training improves model learning to boost performance during testing.

Additionally, in Figure~\ref{fig:contorl}, we demonstrate the capability of Drive-OccWorld to simulate various future occupancies based on specific ego motions, showcasing the potential of Drive-OccWorld as a neural simulator for generating plausible occupancy for autonomous driving.

\begin{table}[t]
\scriptsize
\setlength{\tabcolsep}{2.4pt}
\center
\renewcommand\arraystretch{0.8}
\definecolor{mycolor}{rgb}{0.64, 0.87, 0.93}
\scriptsize
\begin{tabular}{l|cccc|cccc}
\toprule
\multirow{2}{*}{Method}  & \multicolumn{4}{c|}{L2 (m) $\downarrow$} & \multicolumn{4}{c}{Collision (\%) $\downarrow$} \\
& {1s} & {2s} & {3s} & \cellcolor{gray!10}{Avg.} & {1s} & {2s} & {3s} & \cellcolor{gray!10}{Avg.} \\
\midrule
NMP~\cite{zeng2019end}  & - & - & 2.31 & \cellcolor{gray!10}- & - & - & 1.92 & \cellcolor{gray!10}- \\
SA-NMP~\cite{zeng2019end} & - & - & 2.05 & \cellcolor{gray!10}- & - & - & 1.59 & \cellcolor{gray!10}- \\
FF~\cite{hu2021safe}  & 0.55 & 1.20 & 2.54 & \cellcolor{gray!10}1.43 & 0.06 & 0.17 & 1.07 & \cellcolor{gray!10}0.43 \\
EO~\cite{khurana2022differentiable} & 0.67 & 1.36 & 2.78 & \cellcolor{gray!10}1.60 & 0.04 & 0.09 & 0.88 & \cellcolor{gray!10}0.33 \\
\midrule
ST-P3$^{\dagger}$~\cite{hu2022st} & 1.72 & 3.26 & 4.86 & \cellcolor{gray!10}3.28 & 0.44 & 1.08 & 3.01 & \cellcolor{gray!10}1.51  \\
UniAD$^{\dagger}$~\cite{hu2023planning} & 0.48 & 0.96 & 1.65 & \cellcolor{gray!10}1.03 & {0.05} & {0.17} & 0.71 & \cellcolor{gray!10}0.31 \\
VAD-Base$^{\dagger}$~\cite{jiang2023vad} & {0.54} & {1.15} & {1.98} & \cellcolor{gray!10}{1.22} & 0.10 & {0.24} & {0.96} & \cellcolor{gray!10}{0.43} \\
OccNet$^{\dagger}$~\cite{tong2023scene} & 1.29 &2.13 &2.99 &\cellcolor{gray!10}{2.14} & 0.21 & 0.59 &1.37 &\cellcolor{gray!10}{0.72} \\
\rowcolor{mycolor!25} Drive-OccWorld$^{\mathcal{P} \dagger}$ (Ours) & \textbf{0.32} & \textbf{0.75} & \textbf{1.49} & \textbf{0.85} & \textbf{0.05} & \textbf{0.17} & \textbf{0.64} & \textbf{0.29} \\
\midrule
\midrule
ST-P3$^{\ddagger}$~\cite{hu2022st}  & 1.33 & 2.11 & 2.90 & \cellcolor{gray!10}2.11 & 0.23 & 0.62 & 1.27 & \cellcolor{gray!10}0.71 \\
UniAD$^{\ddagger}$~\cite{hu2023planning} & 0.44 & 0.67 & 0.96 & \cellcolor{gray!10}0.69 & {0.04} & {0.08} & 0.23 & \cellcolor{gray!10}0.12  \\
VAD-Base$^{\ddagger}$~\cite{jiang2023vad} & {0.41} & {0.70} & {1.05} & \cellcolor{gray!10}{0.72} & 0.07 & {0.17} & {0.41} & \cellcolor{gray!10}{0.22} \\
Drive-WM$^{\ddagger}$~\cite{wang2024driving} & 0.43 & 0.77 & 1.20 &\cellcolor{gray!10}{0.80} & 0.10 & 0.21 & 0.48 &\cellcolor{gray!10}{0.26} \\
\rowcolor{mycolor!25} Drive-OccWorld$^{\mathcal{P} \ddagger}$ (Ours) & \textbf{0.25} & \textbf{0.44} & \textbf{0.72} & \textbf{0.47} & \textbf{0.03} & \textbf{0.08} & \textbf{0.22} & \textbf{0.11} \\
\midrule
\midrule
UniAD$^{\ddagger *}$~\cite{hu2023planning} & 0.20 & 0.42 & 0.75 &\cellcolor{gray!10}{0.46} & 0.02 & 0.25 & 0.84 &\cellcolor{gray!10}{0.37}  \\
VAD-Base$^{\ddagger *}$~\cite{jiang2023vad} & 0.17 & 0.34 & 0.60 &\cellcolor{gray!10}{0.37} & 0.04 & 0.27 & 0.67 &\cellcolor{gray!10}{0.33} \\
BEV-Planner$^{\ddagger *}$~\cite{li2024ego} & 0.16 & 0.32 & 0.57 &\cellcolor{gray!10}{0.35} & 0.00 & 0.29 & 0.73 & \cellcolor{gray!10}{0.34} \\
\rowcolor{mycolor!25} Drive-OccWorld$^{\mathcal{P} \ddagger *}$ (Ours) & 0.17 & \textbf{0.31} & \textbf{0.49} & \textbf{0.32} & 0.02 & \textbf{0.24} & \textbf{0.62} & \textbf{0.29} \\
\bottomrule
\end{tabular}
\caption{\textbf{End-to-end Planning Performance} on nuScenes. $^{\dagger}$ indicates the $\rm NoAvg$ evaluation protocol, while $^{\ddagger}$ denotes the $\rm TemAvg$ protocol. $^{*}$ signifies the use of ego status in the planning module and the calculations of collision rates following BEV-Planner~\cite{li2024ego}.}
\label{tab4}
\vspace{-1em}
\end{table}

\begin{table}[t] 
\scriptsize
\setlength{\tabcolsep}{2.4pt}
\center
\renewcommand\arraystretch{0.8}
\definecolor{mycolor}{rgb}{0.64, 0.87, 0.93}
\begin{tabular}{ccc|ccc|c}
\toprule
\multicolumn{3}{c|}{Conditional Normalization} & \multirow{2}{*}{$\text{mIoU}_c$} & \multirow{2}{*}{$\text{mIoU}_f$ (1\,s)} & \multirow{2}{*}{$\tilde{\text{mIoU}}_f$} & \multirow{2}{*}{$\text{VPQ}_f^*$} \\
semantic   & ego-motion   & agent-motion  & & & &    \\ \cmidrule{1-7}
 & & & 28.7 & 26.4 & 26.8 & 33.5 \\
\cmark & & & 29.0\textcolor{red}{$_{\uparrow\text{0.3}}$} & 26.6\textcolor{red}{$_{\uparrow\text{0.2}}$} & 27.0\textcolor{red}{$_{\uparrow\text{0.2}}$} & 33.2 \\
 & \cmark &  & 29.4\textcolor{red}{$_{\uparrow\textbf{0.7}}$} & 28.3\textcolor{red}{$_{\uparrow\textbf{1.9}}$} & 28.5\textcolor{red}{$_{\uparrow\textbf{1.7}}$} & 32.6 \\
 & & \cmark & 29.3\textcolor{red}{$_{\uparrow\textbf{0.6}}$} & 27.1\textcolor{red}{$_{\uparrow\text{0.7}}$} & 27.5\textcolor{red}{$_{\uparrow\text{0.7}}$} & 34.4\textcolor{red}{$_{\uparrow\textbf{0.9}}$} \\
\cmidrule{1-7}
\rowcolor{mycolor!25} \cmark & \cmark & \cmark & \textbf{29.4}\textcolor{red}{$_{\uparrow\textbf{0.7}}$} & \textbf{28.3}\textcolor{red}{$_{\uparrow\textbf{1.9}}$} & \textbf{28.6}\textcolor{red}{$_{\uparrow\textbf{1.8}}$} & \textbf{34.5}\textcolor{red}{$_{\uparrow\textbf{1.0}}$} \\
\bottomrule
\end{tabular}
\vspace{-.5em}
\caption{Ablations on the conditional normalization.}
\label{tab5}
\end{table}

\begin{table}[t] 
\scriptsize
\setlength{\tabcolsep}{3.2pt}
\center
\renewcommand\arraystretch{0.8}
\definecolor{mycolor}{rgb}{0.64, 0.87, 0.93}
\begin{tabular}{cc|c|ccc|c}
\toprule
\multicolumn{2}{c|}{Condition Interface} & Fourier & \multirow{2}{*}{$\text{mIoU}_c$} & \multirow{2}{*}{$\text{mIoU}_f$ (1\,s)} & \multirow{2}{*}{$\tilde{\text{mIoU}}_f$} & \multirow{2}{*}{$\text{VPQ}_f^*$} \\
addition & cross-attention & Embed  &  & & &    \\ \cmidrule{1-7}
 & & & 28.7 & 26.4 & 26.8 & 33.5 \\
 \cmark & & \cmark & 28.9\textcolor{red}{$_{\uparrow\text{0.2}}$} & 27.4\textcolor{red}{$_{\uparrow\text{1.0}}$} & \textbf{28.0}\textcolor{red}{$_{\uparrow\text{1.2}}$} & 34.2\textcolor{red}{$_{\uparrow\text{0.7}}$} \\
 & \cmark & & 28.5 & 27.1\textcolor{red}{$_{\uparrow\text{0.7}}$} & 27.4\textcolor{red}{$_{\uparrow\text{0.6}}$} & 33.9\textcolor{red}{$_{\uparrow\text{0.4}}$} \\
 \cmidrule{1-7}
\rowcolor{mycolor!25}  & \cmark & \cmark & \textbf{29.0}\textcolor{red}{$_{\uparrow\text{0.3}}$} & \textbf{27.6}\textcolor{red}{$_{\uparrow\text{1.2}}$} & 27.8\textcolor{red}{$_{\uparrow\text{1.0}}$} & \textbf{35.0}\textcolor{red}{$_{\uparrow\text{1.5}}$} \\
\bottomrule
\end{tabular}
\vspace{-.5em}
\caption{Ablations on the action conditioning interface.}
\label{tab6}
\end{table}

\begin{table}[t]
\scriptsize
\setlength{\tabcolsep}{2.8pt}
\center
\renewcommand\arraystretch{0.8}
\definecolor{mycolor}{rgb}{0.64, 0.87, 0.93}
\scriptsize
\begin{tabular}{ccc|c|cccc|cccc}
\toprule
\multicolumn{3}{c|}{Cost Factors}  & BEV & \multicolumn{4}{c|}{L2 (m) $\downarrow$} & \multicolumn{4}{c}{Collision (\%) $\downarrow$} \\
Agent & Road & Volume & Refine & {0.5s} & {1s} & {1.5s} & \cellcolor{gray!10}{Avg.} & {0.5s} & {1s} & {1.5s} & \cellcolor{gray!10}{Avg.} \\
\midrule
\xmark & \cmark & \cmark & \cmark & 0.15 & 0.30 & 0.50 & \cellcolor{gray!10}0.32 & 0.14 & 0.16 & 0.18 & \cellcolor{gray!10}0.16 \\
\cmark & \xmark & \cmark & \cmark & 0.14 & 0.28 & 0.46 & \cellcolor{gray!10}0.29 & 0.09 & 0.11 & 0.13 & \cellcolor{gray!10}0.11 \\
\cmark & \cmark & \xmark & \cmark & 0.14 & 0.27 & \textbf{0.44} & \cellcolor{gray!10}0.28 & 0.09 & 0.14 & 0.18 & \cellcolor{gray!10}0.14 \\
\cmark & \cmark & \cmark & \xmark & 0.22 & 0.36 & 0.52 & \cellcolor{gray!10}0.37 & 0.14 & 0.20 & 0.27 & \cellcolor{gray!10}0.20 \\
\midrule
\rowcolor{mycolor!25} \cmark & \cmark & \cmark & \cmark & \textbf{0.11} & \textbf{0.26} & 0.46 & \textbf{0.28} & \textbf{0.04} & \textbf{0.11} & \textbf{0.13} & \textbf{0.09} \\
\bottomrule
\end{tabular}
\vspace{-.5em}
\caption{Contributions of occupancy-based cost factors.}
\label{tab7}
\end{table}

\subsection{End-to-end Planning with Drive-OccWorld}
Table~\ref{tab4} presents the planning performance compared to existing end-to-end methods in terms of L2 error and collision rate. We provide results under different evaluation protocol settings from ST-P3 and UniAD. Specifically, $\rm NoAvg$ denotes the result at the corresponding timestamp, while $\rm TemAvg$ calculates metrics by averaging performances from 0.5s to the corresponding timestamp.

As shown in Table~\ref{tab4}, Drive-OccWorld$^\mathcal{P}$ achieves superior planning performance compared to existing methods. For instance, Drive-OccWorld$^{\mathcal{P} \dagger}$ obtains relative improvements of 33\%, 22\%, and 9.7\% on L2@1s, L2@2s, and L2@3s, respectively, compared to UniAD$^\dagger$. We attribute this improvement to the world model's capacity to accumulate world knowledge and envision future states. It effectively enhances the planning results for future timestamps and improves the safety and robustness of end-to-end planning.

Recent studies \cite{li2024ego} have examined the impact of incorporating ego status into the planning module. In line with this research, we also conduct a fair comparison between our model equipped with ego status and previous works. Our findings indicate that Drive-OccWorld still achieves the highest performance at distant future timestamps, demonstrating the effectiveness of continuous forecasting and planning.

\subsection{Ablation Study}
The default configuration for the ablation experiments involves using one historical and the current images as input to predict the inflated GMO over two future timestamps.

\noindent\textbf{Conditional Normalization.}
In Table \ref{tab5}, we ablate the conditional normalization method while discarding the action conditions in Sec.~\ref{action} to avoid potential influence.
The results indicate that each pattern yields gains, particularly with ego-motion aware normalization achieving a 1.9\% increase in $\text{mIoU}_{f}$, highlighting the importance of ego status for future state forecasting. Additionally, agent-motion aware normalization enhances $\text{VQP}_{f}^*$ by 0.9\% by compensating for the movements of other agents.

\noindent\textbf{Action Conditioning Interface.}
In Table \ref{tab6}, we investigate the method of injecting action conditions into the world decoder. Compared to adding conditions to BEV queries, cross-attention is a more effective approach for integrating prior knowledge into the generation process. Furthermore, Fourier embedding provides additional improvement by encoding conditions into latent space at high frequencies.

\noindent\textbf{Occupany-based Costs.}
Table \ref{tab7} ablates the occupancy-based cost function, and the results indicate that each cost factor contributes to safe planning, particularly highlighting that the absence of agent constraints results in a higher collision rate. Additionally, BEV refinement is vital as it provides more comprehensive 3D information about the environment.

\section{Conclusion}
We propose Drive-OccWorld, a 4D occupancy forecasting and planning world model for autonomous driving. Flexible action conditions can be injected into the world model for action-controllable generation, facilitating a broader range of downstream applications. An occupancy-based planner is integrated with the world model for motion planning, considering both safety and the 3D structure of the environment. Experiments demonstrate that our method exhibits remarkable performance in occupancy and flow forecasting. Planning results are improved by leveraging the world model's capacity to accumulate world knowledge and envision future states, thereby enhancing the safety and robustness of end-to-end planning.

\section*{Acknowledgments}
We thank Jiangning Zhang and Jiang He for helpful discussions and valuable support on the paper. We thank all authors for their contributions.
This work was supported by a Grant from The National Natural Science Foundation of China (No. 62103363).

\bibliography{aaai25}

\twocolumn[
    \centering
    \textbf{\Large Driving in the Occupancy World: Vision-Centric 4D Occupancy Forecasting \\ and Planning via World Models for Autonomous Driving}\\
    \vspace{1.2em}
    \Large Supplementary Material
    \vspace{1.2em}
    ]

\appendix

\hypersetup{linkcolor=black}
\tableofcontents
\hypersetup{linkcolor=red}
\addtocontents{toc}{\protect\setcounter{tocdepth}{2}}

\section{Additional Related Works} \label{related work}
\subsection{End-to-end Autonomous Driving}  \label{e2eAD}
In recent decades, autonomous driving (AD) algorithms~\cite{ayoub2019manual, chen2023milestones, mei2023centerlps} have significantly advanced from modular pipelines~\cite{guo2023scenedm, li2023logonet} to end-to-end models~\cite{hu2023planning, jiang2023vad}, which predict planning trajectories directly from raw sensor data in a unified manner. 
For instance, P3 \cite{sadat2020perceive} and ST-P3 \cite{hu2022st} learn a differentiable occupancy representation from the perception module, serving as a cost factor for producing safe maneuvers. Building on the success of detection \cite{huang2021bevdet, li2023bevstereo, li2023bevdepth, li2022bevformer, wang2023exploring, li2024bevnext} and segmentation \cite{ng2020bev, peng2023bevsegformer} in Bird's-eye view (BEV), several methods plan trajectories based on BEV perception and prediction.
UniAD \cite{hu2023planning} is a representative planning-oriented end-to-end model that integrates BEV tracking and motion prediction with planning. VAD \cite{jiang2023vad} employs a vectorized representation in BEV for scene learning and planning. GraphAD \cite{zhang2024graphad} utilizes a graph model to establish interactions among agents and map elements. DualAD \cite{doll2024dualad} disentangles dynamic and static representations using object and BEV queries to address the effects of object motion. UAD \cite{guo2024end} proposes an unsupervised proxy to reduce the need for costly annotations across multiple modules. PARA-Drive \cite{weng2024drive} explores module connectivity and introduces a fully parallel architecture. SparseAD~\cite{zhang2024sparsead} and SparseDrive~\cite{sun2024sparsedrive} investigate sparse representations to enhance the efficacy and efficiency of end-to-end methods.

\subsection{Occupancy Prediction and Forecasting}   \label{occ}
\paragraph{Occupancy Prediction} aims to construct the occupancy state of the surrounding environment based on observations. Existing methods are primarily classified into LiDAR-based and camera-based approaches, depending on the input modality.

LiDAR-based methods \cite{song2017semantic, roldao2020lmscnet, yan2021sparse, yang2021semantic, mei2023ssc} derive occupancy predictions primarily from grids generated by sparse LiDAR points, investigating the relationships between semantic segmentation and scene completion. In contrast, camera-based methods \cite{cao2022monoscene, li2023voxformer, wei2023surroundocc, wang2023openoccupancy, mei2023camera} focus on efficiently transforming 2D image features into 3D representations. For example, TPVFormer \cite{huang2023tri} introduces a tri-perspective view (TPV) representation to describe the 3D structure, OccFormer \cite{zhang2023occformer} employs depth prediction for image-to-3D transformation, and Occ3D \cite{tian2023occ3d} utilizes cross-attention to aggregate 2D features into 3D space. 
Recent methods \cite{huang2024selfocc, zhang2023occnerf, boeder2024occflownet} also implement occupancy prediction in a self-supervised manner, utilizing neural rendering to convert occupancy fields into depth maps, which can then be supervised using multi-frame photometric consistency.

\paragraph{Occupancy Forecasting} aims to predict the near-future occupancy state (\emph{i.e.}, how the surroundings evolve) based on historical and current observations. Existing methods \cite{khurana2022differentiable, khurana2023point, toyungyernsub2022dynamics} primarily utilize LiDAR point clouds to identify changes in the surrounding structure and predict future states. For example, Khurana et al. \cite{khurana2022differentiable} propose a differentiable depth rendering method that generates point clouds from 4D occupancy predictions, facilitating occupancy forecasting training with unannotated LiDAR sequences. Other point cloud prediction methods \cite{lu2021monet, mersch2022self, luo2023pcpnet} directly forecast future laser points, which can then be voxelized for subsequent occupancy estimation. Cam4DOcc \cite{ma2024cam4docc} is the first framework for camera-based 4D occupancy forecasting, establishing a benchmark that provides sequential occupancy states for both movable and static objects, along with their 3D backward centripetal flow.

The aforementioned methods can only predict future occupancy states using historical observations. In this work, we propose a framework that simultaneously predicts future occupancy and generates planning trajectories based on the occupancy status. Additionally, we implement action-conditioned occupancy generation to associate future predictions with various ego actions.

\begin{figure}[t]
\centering
    \setlength{\abovecaptionskip}{-.2em}
    \setlength{\belowcaptionskip}{-1.7em}
	\includegraphics[width=0.98\linewidth]{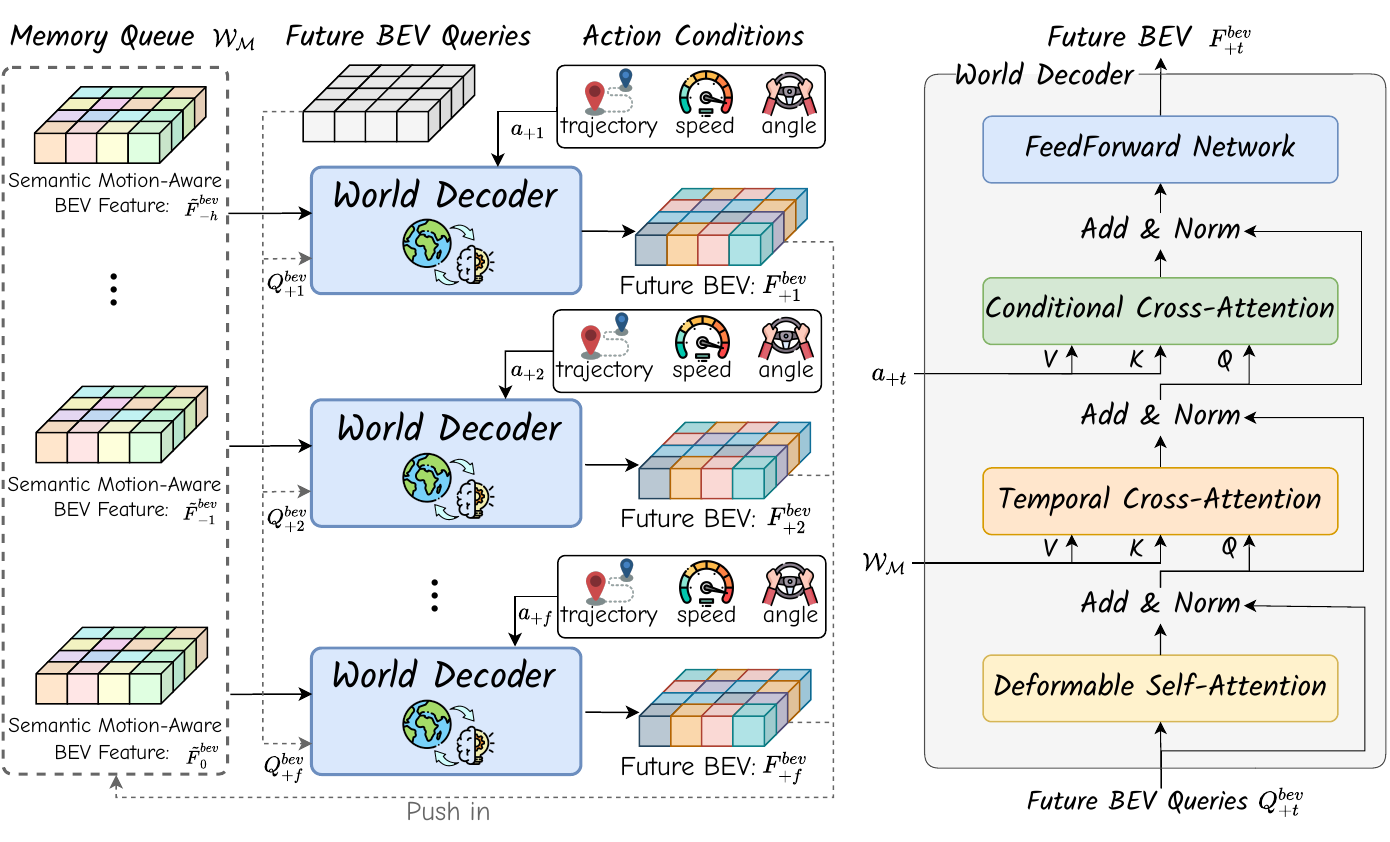}
	\caption{\textbf{Detailed structure of the world decoder}, which predicts the next BEV features based on historical BEV features and expected ego actions in an autoregressive manner.}
	\label{fig:decoder}
\end{figure}

\section{Additional Method Details} \label{method}
\subsection{World Decoder}  \label{decoder}
As depicted in Figure~\ref{fig:decoder}, the world decoder $\mathcal{W}_\mathcal{D}$ predicts the BEV embeddings at the next timestamp based on historical BEV features stored in $\mathcal{W}_\mathcal{M}$ and expected action conditions in an autoregressive manner.

Specifically, the learnable BEV queries $Q \in \mathbb{R}^{h \times w \times c}$ first perform deformable self-attention to establish contextual relationships. A temporal cross-attention layer is then employed to extract corresponding features from multi-frame historical embeddings. To save memory while maintaining efficiency, this layer also operates in a deformable manner. Notably, the movement of the ego vehicle can cause misalignment between BEV embeddings at different timestamps; thus, reference coordinates are calculated using ego transformations for feature alignment. Subsequently, a conditional cross-attention layer is utilized to perform cross-attention between the BEV queries and conditional embeddings, injecting action conditions into the forecasting process. Finally, a feedforward network outputs the generated BEV features, which can be used for future occupancy and flow forecasting.

\subsection{Loss Function}  \label{loss}
We optimize Drive-OccWorld with historical normalization, future forecasting, and trajectory planning in an end-to-end manner by leveraging the following loss functions:
\begin{equation}
    \mathcal{L} = \mathcal{L}_{norm} + \mathcal{L}_{fcst} + \mathcal{L}_{plan}
\end{equation}

\paragraph{Normalization Loss.} In semantic-conditional normalization, we employ cross-entropy loss to supervise the historical semantic probabilities $\mathcal{S}_{-h:0}$, thereby learning effective semantics for affine transformation. Additionally, $l_1$ loss is used to supervise the historical 3D backward centripetal flow predictions $\mathcal{F}_{-h:0}$. These losses are combined to form the normalization loss $\mathcal{L}_{norm}$, which enhances both semantic and dynamic representations in the historical memories.

\paragraph{Forecasting Loss.} For the future forecasting loss $\mathcal{L}_{fcst}$, flow predictions $\mathcal{F}_{1:f}$ are supervised using $l_1$ loss, while occupancy predictions $\mathcal{S}_{1:f}$ are constrained by multiple losses, including cross-entropy loss $\mathcal{L}_{ce}$, Lovász loss $\mathcal{L}_{lovasz}$, and binary occupancy loss $\mathcal{L}_{bce}$:
\begin{equation}
    \mathcal{L}_{occ} = \frac{1}{N_f}\sum_{t=1}^{N_f} (\mathcal{L}_{ce}(\mathcal{S}_t, \hat{\mathcal{S}_t}) + \mathcal{L}_{lovasz}(\mathcal{S}_t, \hat{\mathcal{S}}_t) + \mathcal{L}_{bce}(\mathcal{O}_t, \hat{\mathcal{O}_t}))
\end{equation}
where $N_f$ denotes the number of future frames, $\hat{\mathcal{S}}_t$ represents the ground-truth semantic occupancy at timestamp $t$, and $\mathcal{O}$ is the binary occupancy. These losses are combined to constrain the semantics and geometries of occupancy forecasting, guiding Drive-OccWorld in learning how the world evolves.

\paragraph{Planning Loss.}
Since selecting the lowest-cost trajectory from a discrete set $\tau^*$ is not differentiable, we employ a max-margin loss inspired by ST-P3 \cite{sadat2020perceive, hu2022st} to penalize trajectory candidates with low costs that deviate from the expert trajectory. Let $o$ represent the occupied grids by agents and roads, and let $\hat{\tau}$ denote the expert trajectory. The max-margin loss encourages the expert trajectory to have a smaller occupancy cost than other candidates.

Overall, the planning loss consists of three components: the max-margin loss, a naive $l_2$ loss for imitation learning, and a collision loss $l_{coll}$ that ensures the planned trajectory avoids grids occupied by obstacles:
\begin{equation}
    \mathcal{L}_{plan} = \underset{\tau^*}{\text{max}}[f_o(o, \hat{\tau}) - f_o(o, \tau^*)]_+ + l_2(\tau_o, \hat{\tau}) + l_{coll}(\tau_o, a)
\end{equation}
where $f_o$ is the occupancy-based cost function, $\tau_o$ denotes the final predicted trajectory, $a$ represents the grids occupied by obstacles, and $[\cdot]_+$ denotes the ReLU function.

\section{Experimental Settings} \label{experiment}
\subsection{Dataset} \label{dataset}
We adopt the nuScenes \cite{caesar2020nuscenes}, nuScenes-Occupancy \cite{wang2023openoccupancy}, and Lyft-Level5 datasets for occupancy forecasting and planning.
Following Cam4DOcc \cite{ma2024cam4docc}, we use the corresponding images from two historical frames and one current frame as input to predict future states over four timestamps. Since nuScenes is annotated at 2Hz while Lyft-Level5 is annotated at 5Hz, we report the results with 2s and 0.8s time intervals, respectively.
After reorganizing the data, the number of sequences for training and testing is 23,930 and 5,119 in scenes and 15,720 and 5,880 in Lyft-Level5.

\subsection{Metrics} \label{metirc}
\paragraph{Occupancy Forecasting} is evaluated using the mIoU metric, with $\text{mIoU}_c$ and $\text{mIoU}_f$ denoting the performances of the current and future timestamps, respectively:
\begin{equation}
    \text{mIoU}_c(\mathcal{S}_c, \hat{\mathcal{S}}_c) = \frac{\sum_{H,W,L} \mathcal{S}_c \cdot \hat{\mathcal{S}}_c}{\sum_{H,W,L} \mathcal{S}_c + \hat{\mathcal{S}}_c - \mathcal{S}_c \cdot \hat{\mathcal{S}}_c}
\end{equation}
\begin{equation}
    \text{mIoU}_f(\mathcal{S}_f, \hat{\mathcal{S}}_f) = \frac{1}{N_f} \sum_{t=1}^{N_f} \frac{\sum_{H,W,L} \mathcal{S}_t \cdot \hat{\mathcal{S}}_t} {\sum_{H,W,L} \mathcal{S}_t + \hat{\mathcal{S}}_t - \mathcal{S}_t \cdot \hat{\mathcal{S}}_t}
\end{equation}
where $\mathcal{S}_t$ and $\hat{\mathcal{S}}_t$ represent the predicted and ground-truth semantic occupancy at timestamp $t$, respectively. Here, $H$, $W$, and $L$ denote the dimensions of the 3D volume, while $N_f$ represents the number of future frames.

Additionally, we introduce a quantitative indicator $\tilde{\text{mIoU}}_f$, weighted by timestamp, based on the principle that occupancy predictions at nearby timestamps are more critical for planning:
\begin{equation}
    \tilde{\text{mIoU}}_f(\mathcal{S}_f, \hat{\mathcal{S}}_f) = \frac{1}{N_f} \sum_{t=1}^{N_f} \frac{1}{t} \sum_{k=1}^t \frac{\sum_{H,W,L} \mathcal{S}_k \cdot \hat{\mathcal{S}}_k}{\sum_{H,W,L} \mathcal{S}_k + \hat{\mathcal{S}}_k - \mathcal{S}_k \cdot \hat{\mathcal{S}}_k}
\end{equation}

\paragraph{Flow Forecasting} is evaluated by extending the video panoptic quality (VPQ) metric~\cite{kim2020video} from 2D instance prediction to 3D instance prediction. Specifically, instance centers are determined from occupancy predictions at $t=0$ using non-maximum suppression (NMS), and voxel-wise instance IDs are assigned using the predicted 3D backward centripetal flow over $t \in [1,f]$. Since instance centers may be duplicated during the NMS operation, we introduce a simple yet efficient center clustering technique that combines the center proposals based on their relative distances, merging them into a more unified set. The $\text{VPQ}_f$ can be calculated as follows:
\begin{equation}
    \text{VPQ}_f(\mathcal{F}_f, \hat{\mathcal{F}}_f) = \frac{1}{N_f} \sum_{t=0}^{N_f} \frac{\sum_{(p_t, q_t) \in TP_t} \text{IoU}(p_t, q_t)}{|TP_t| + \frac{1}{2}|FP_t| + \frac{1}{2}|FN_t|}
\end{equation}
where $|TP_t|$, $|FP_t|$, and $|FN_t|$ represent the true positives, false positives, and false negatives at timestamp $t$, respectively. Following Cam4DOcc, a predicted instance is considered a true positive if its $\text{IoU}$ is greater than 0.2 and the corresponding instance ID is correctly tracked.

\paragraph{Planning} is evaluated using L2 error and collision rate at various timestamps. We also report the modified collision rate proposed in~\cite{li2024ego}, where the collisions at each moment depend on previous timestamps:
\begin{equation}
    CR(t) = \sum_{t=0}^{N_f} \mathbb{I}_t > 0
\end{equation}
where $\mathbb{I}_t$ indicates whether the ego vehicle at timestamp $t$ will intersect with other obstacles.

\subsection{Implementation Details} \label{implementation}
The 3D environment is voxelized within $[[\pm51.2 \text{m}], [\pm51.2 \text{m}], [-5 \text{m}, 3\text{m}]]$ at a voxel resolution of $0.2 \text{m}$, resulting in a volume size of $512 \times 512 \times 40$. The historical encoder $\mathcal{W}_\mathcal{E}$ is derived from the BEVFormer-based encoder \cite{li2022bevformer}, which includes an image backbone with a Feature Pyramid Network (FPN) neck and six additional transformer encoder layers. The spatial resolutions of BEV queries are set to $h, w = 200$. The world decoder $\mathcal{W}_\mathcal{D}$ consists of three layers, each with 256 channels. The prediction heads have output channels set to 16 to convert BEV embeddings into 3D semantic occupancy and flow predictions $(\mathcal{S}, \mathcal{F}) \in \mathbb{R}^{200 \times 200 \times 16}$. During training, we use historical multi-view camera images over two timesteps to forecast future occupancy over two seconds (with each timestep corresponding to a 0.5-second interval). We employ an AdamW optimizer with an initial learning rate of $2 \times 10^{-4}$ and a cosine annealing scheduler to train Drive-OccWorld on 8 NVIDIA A100 GPUs.

\section{Additional Experiments}

\begin{table}[t] 
\scriptsize
\setlength{\tabcolsep}{2.8pt}
\center
\renewcommand\arraystretch{0.9}
\definecolor{mycolor}{rgb}{0.64, 0.87, 0.93}
\begin{tabular}{cc|c|ccc|ccc}
\toprule
\multicolumn{2}{c|}{No. of Frames} & Memory & \multirow{2}{*}{$\text{mIoU}_c$} & \multirow{2}{*}{$\text{mIoU}_f$ (2\,s)} & \multirow{2}{*}{$\tilde{\text{mIoU}}_f$} & \multicolumn{3}{c}{Latency (ms)} \\
History & Current & Length & & & & $\mathcal{W}_\mathcal{E}$ & $\mathcal{W}_\mathcal{M}$ & $\mathcal{W}_\mathcal{D}$   \\ \cmidrule{1-9}
 \cmidrule{1-7}
0 & 1 & 1 & 13.0 & 12.1 & 12.3 & 140 & 10 & 190 \\
1 & 1 & 1 & 14.3\textcolor{red}{$_{\uparrow\text{1.3}}$} & 13.4\textcolor{red}{$_{\uparrow\text{1.3}}$} & 13.6\textcolor{red}{$_{\uparrow\text{1.3}}$} & 269 & 12 & 188 \\
2 & 1 & 2 & 14.8\textcolor{red}{$_{\uparrow\text{1.8}}$} & 13.8\textcolor{red}{$_{\uparrow\text{1.7}}$} & 14.1\textcolor{red}{$_{\uparrow\text{1.8}}$} & 406 & 19 & 207 \\
\rowcolor{mycolor!25} 2 & 1 & 3 & 15.1\textcolor{red}{$_{\uparrow\text{2.1}}$} & 14.1\textcolor{red}{$_{\uparrow\text{2.0}}$} & 14.3\textcolor{red}{$_{\uparrow\text{2.0}}$} & 401 & 26 & 220 \\
\bottomrule
\end{tabular}
\captionsetup{aboveskip=0.6em}
\setlength{\belowcaptionskip}{-0.5em}
\caption{\textbf{Latency and performance} based on varying numbers of input frames and memory queue lengths. Latency measurements are conducted on an A6000 GPU.}
\label{tab:efficiency}
\end{table}

\begin{table}[t] 
\scriptsize
\setlength{\tabcolsep}{1.8pt}
\center
\renewcommand\arraystretch{0.8}
\definecolor{mycolor}{rgb}{0.64, 0.87, 0.93}
\begin{tabular}{ccc|ccc|c}
\toprule
\multicolumn{3}{c|}{Semantic Loss} & \multirow{2}{*}{$\text{mIoU}_c$} & \multirow{2}{*}{$\text{mIoU}_f$ (1\,s)} & \multirow{2}{*}{$\tilde{\text{mIoU}}_f$} & \multirow{2}{*}{$\text{VPQ}_f^*$} \\
Cross Entropy & Binary Occupancy  & Lovasz  & & &   \\ 
\cmidrule{1-7}
\cmark & & & 28.3 & 27.2 & 27.5 & 34.1 \\
\cmark & \cmark & & 28.7\textcolor{red}{$_{\uparrow\text{0.4}}$} & 27.9\textcolor{red}{$_{\uparrow\text{0.7}}$} & 28.1\textcolor{red}{$_{\uparrow\text{0.6}}$} & 34.3\textcolor{red}{$_{\uparrow\text{0.2}}$} \\
\rowcolor{mycolor!25} \cmark & \cmark & \cmark & \textbf{29.4}\textcolor{red}{$_{\uparrow\textbf{1.1}}$} & \textbf{28.3}\textcolor{red}{$_{\uparrow\textbf{1.1}}$} & \textbf{28.6}\textcolor{red}{$_{\uparrow\textbf{1.1}}$} & 34.5\textcolor{red}{$_{\uparrow\text{0.4}}$}\\
\bottomrule
\end{tabular}
\captionsetup{aboveskip=1.0em}
\setlength{\belowcaptionskip}{-2em}
\caption{\textbf{Ablation studies on the semantic occupancy loss functions}, utilizing one historical and the current inputs to predict future states across two timestamps.}
\label{tab:sem_loss}
\end{table}

\begin{figure*}[t]
\centering
	\includegraphics[width=0.95\textwidth]{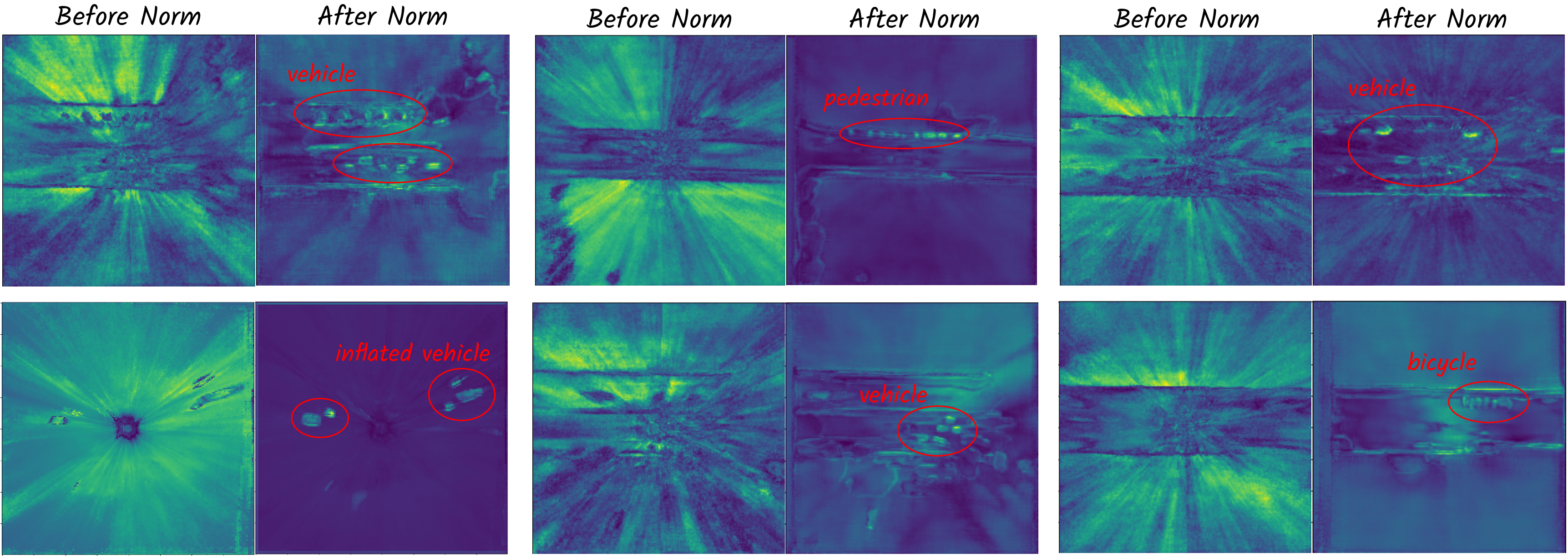}
    \setlength{\abovecaptionskip}{0.5em}
    \setlength{\belowcaptionskip}{-0.5em}
	\caption{\textbf{Visualization of BEV features before and after semantic-conditional normalization} highlights the responses of BEV grids, particularly for instance objects. Consequently, it extracts discriminative BEV features for forecasting and planning.}
	\label{fig:sem_norm}
\end{figure*}

\begin{figure*}[t]
\centering
	\includegraphics[width=0.95\textwidth]{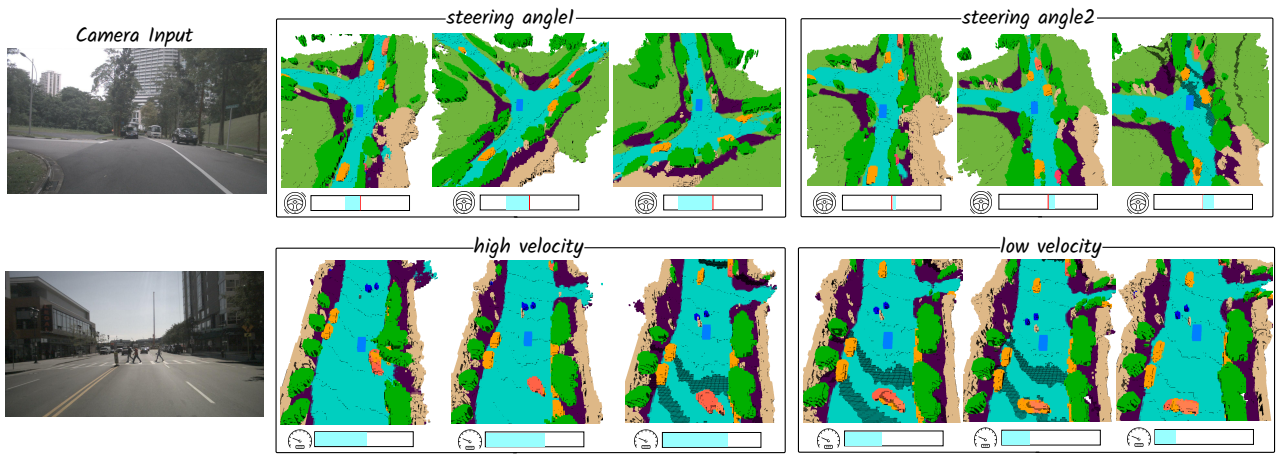}
    \setlength{\abovecaptionskip}{0.5em}
    \setlength{\belowcaptionskip}{-0.5em}
	\caption{\textbf{Qualitative results of controllable generation}, using the high-level \emph{steering angle} or low-level \emph{velocity} conditions.}
	\label{fig:control2}
\end{figure*}

\subsection{Effectiveness and Efficiency} \label{memory}
In Table~\ref{tab:efficiency}, we compare the performance and latency of different configurations based on the number of input frames and the length of the memory queue. We observe that without historical inputs, the performance at the current moment, $\text{mIoU}_c$, is slightly weaker. However, it can still generate plausible future forecasting results while operating at real-time speed. Furthermore, increasing both the number of input frames and the length of the memory queue leads to improvements in performance.

Regarding latency, most of the time consumption is attributed to the historical encoder $\mathcal{W}_\mathcal{E}$. In contrast, the memory queue $\mathcal{W}_\mathcal{M}$ and the world decoder $\mathcal{W}_\mathcal{D}$ exhibit limited latency, even as the number of input frames increases, demonstrating the efficiency of our architectural design. We further test the latencies of the modules in $\mathcal{W}_\mathcal{E}$, where the image backbone runs at 27 ms, the FPN neck runs at 13 ms, and the transformer encoder runs at 100 ms. In future work, we will explore sparse representations such as SparseBEV~\cite{liu2023sparsebev} to enhance the efficacy and efficiency of Drive-OccWorld.

\subsection{Semantic-Conditional Normalization} \label{semantic norm}
The original BEV embeddings are derived from 2D image features and exhibit ray-shaped patterns, as shown in the visualizations of \emph{Before Normalization} in Figure~\ref{fig:sem_norm}. This phenomenon can be explained by the fact that waypoints along the same ray in 3D space often correspond to the same pixel in the image, leading to the extraction of similar feature representations. Consequently, these ray-shaped features are not sufficiently discriminative for semantic occupancy predictions.

To address this issue, we propose semantic-conditional normalization, which emphasizes BEV features with higher semantic-conditional probabilities, making the BEV embeddings more semantically discriminative. As illustrated in the visualizations of BEV features \emph{After Normalization} in Figure~\ref{fig:sem_norm}, this approach efficiently highlights the responses of BEV grids, particularly for instance objects such as vehicles, pedestrians, and bicycles, which are crucial for occupancy prediction and collision avoidance. Meanwhile, it preserves both semantic and geometric information, benefiting subsequent semantic occupancy forecasting and planning.

\subsection{Semantic Loss} \label{semantic loss}
In Table~\ref{tab:sem_loss}, we ablate the effectiveness of various loss functions for semantic occupancy forecasting, including cross-entropy loss, binary occupancy loss, and Lovász loss. The results reveal that using only cross-entropy loss achieves satisfactory performance while adding additional supervision leads to further gains. Interestingly, semantic losses also enhance the performance of flow forecasting on $\text{VPQ}_f^*$, demonstrating that accurate object center localization is crucial for instance association.

\section{Additional Visualizations}

\subsection{Action Controllability} \label{control}
In Figure~\ref{fig:control2}, we present additional qualitative results of the controllable generation of Drive-OccWorld, which takes historical visual images and diverse action conditions as input to predict various future states. The results illustrate that Drive-OccWorld can generate plausible future occupancies based on specific ego actions. For instance, the bottom section showcases the generated occupancies using different ego velocities, where high velocity brings the ego vehicle dangerously close to colliding with pedestrians. These qualitative results demonstrate the potential of Drive-OccWorld as a neural simulator to generate realistic environmental states for autonomous driving, facilitating a broader range of downstream applications.

\begin{figure*}[htbp]
\centering
	\includegraphics[width=0.98\textwidth]{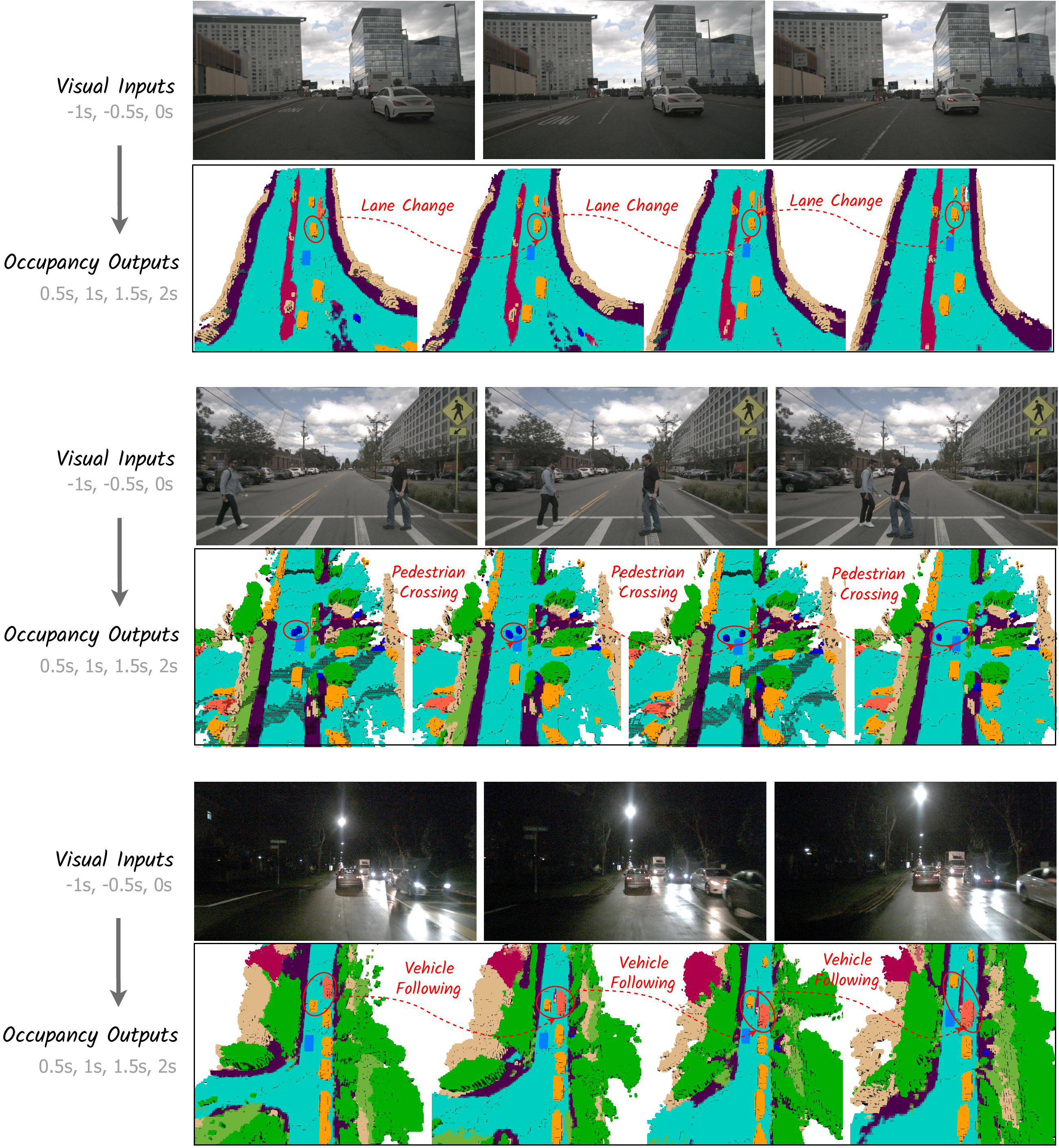}
    \caption{\textbf{Qualitative results for vision-centric 4D occupancy forecasting on the nuScenes validation set}. Top: historical visual inputs over three timestamps, showing only the front images for simplicity. Bottom: future occupancy predictions for two seconds. Notable movable objects are highlighted in \textcolor{red}{red circles}. The first section indicates a vehicle changing lanes, the second section shows two pedestrians crossing the road, and the third section depicts the ego vehicle following a preceding car.}
	\label{fig:forecasting}
\end{figure*}

\begin{figure*}[htbp]
\centering
	\includegraphics[width=0.98\textwidth]{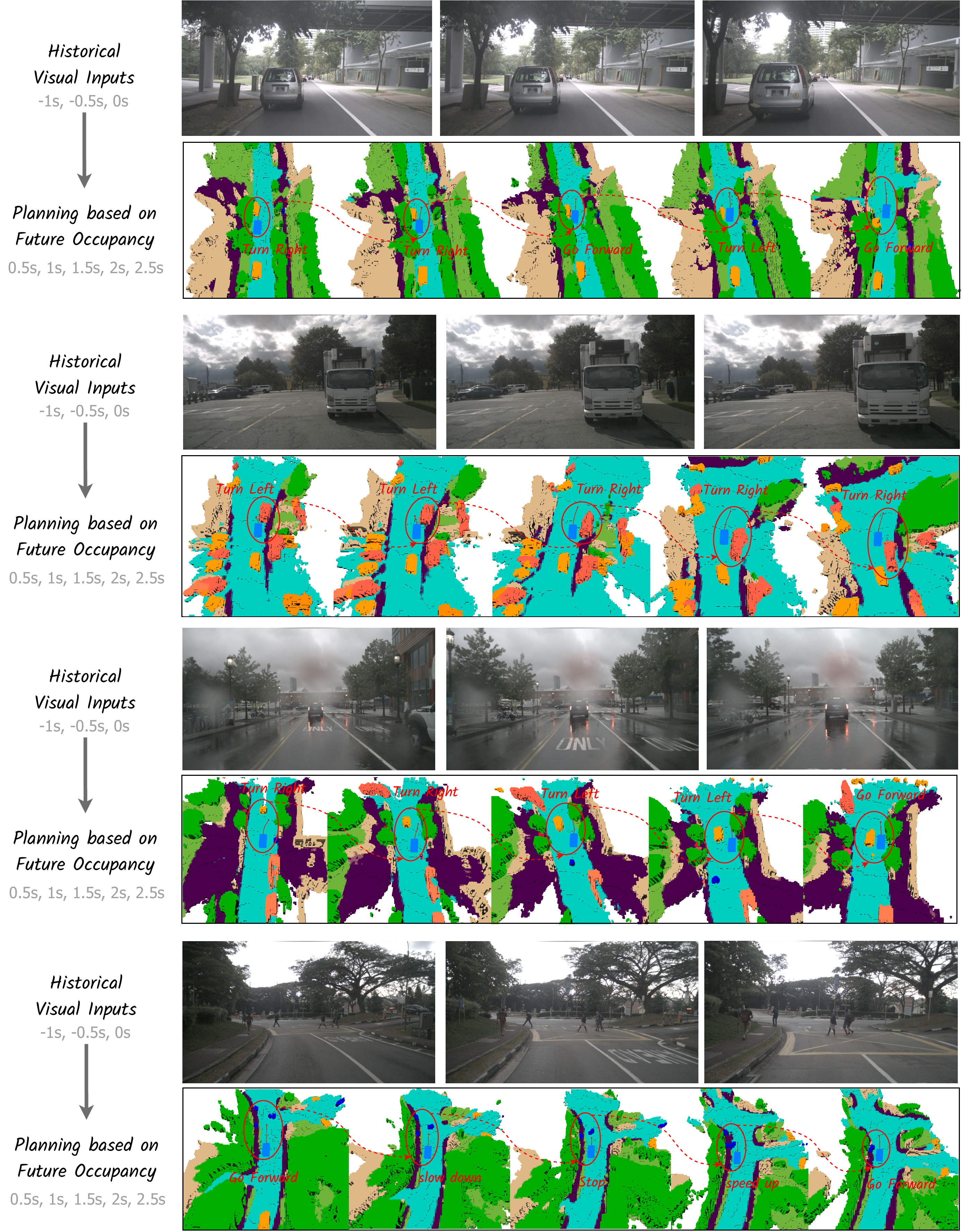}
    \caption{\textbf{Qualitative results for continuous forecasting and planning on the nuScenes validation set.} Top: historical visual inputs over three timestamps, showing only the front images for simplicity. Bottom: future occupancy and planning results for five timestamps, with predicted trajectories highlighted in \textcolor{red}{red arrows}. The first three sections show the ego vehicle avoiding preceding cars, while the last section illustrates the ego car slowing down and stopping until the pedestrians in front have passed.}
	\label{fig:plan}
\end{figure*}

\subsection{4D Occupancy Forecasting}
In Figure~\ref{fig:forecasting}, we present additional qualitative results for vision-centric 4D occupancy forecasting. The historical visual inputs over three timestamps are displayed in the top part, while the future occupancy predictions for two seconds are shown in the bottom part. Notable movable objects around the ego vehicle are highlighted with \textcolor{red}{red circles}.

The first section depicts a vehicle changing its driving lane from right to left, with Drive-OccWorld accurately predicting its future states. The second section illustrates two pedestrians crossing the road in front of the ego vehicle, and Drive-OccWorld successfully predicts their future trajectories. The third section reveals that Drive-OccWorld learns that the ego vehicle follows a preceding car, even under nighttime conditions. These results demonstrate that Drive-OccWorld understands how the world evolves by accurately modeling the dynamics of movable objects and the future states of the static environment.

\subsection{Trajectory Planning}   \label{traj}
In Figure~\ref{fig:plan}, we present qualitative results of trajectory planning based on occupancy predictions. The predicted trajectories are highlighted with \textcolor{red}{red arrows}.

The first three sections demonstrate how the ego vehicle avoids preceding cars by turning, showcasing the effectiveness of the agent-safety cost in preventing collisions. The third section also indicates that Drive-OccWorld is adaptable to rainy conditions. The last section illustrates the ego vehicle slowing down and stopping to allow pedestrians in front to cross the road before resuming speed for future navigation.

\end{document}